*Research Article*

# RLAS-BIABC: A Reinforcement Learning-Based Answer Selection Using the BERT Model Boosted by an Improved ABC Algorithm

**Hamid Gharagozlou**,[1] **Javad Mohammadzadeh**,[1] **Azam Bastanfard**,[1] **and Saeed Shiry Ghidary**[2]

[1]*Department of Computer Engineering, Karaj Branch, Islamic Azad University, Karaj, Iran*
[2]*School of Digital, Technologies, and Arts, Staffordshire University, Stoke-on-Trent, UK*

Correspondence should be addressed to Javad Mohammadzadeh; j.mohammadzadeh@kiau.ac.ir

Answer selection (AS) is a critical subtask of the open-domain question answering (QA) problem. The present paper proposes a method called RLAS-BIABC for AS, which is established on attention mechanism-based long short-term memory (LSTM) and the bidirectional encoder representations from transformers (BERT) word embedding, enriched by an improved artificial bee colony (ABC) algorithm for pretraining and a reinforcement learning-based algorithm for training backpropagation (BP) algorithm. BERT can be comprised in downstream work and fine-tuned as a united task-specific architecture, and the pretrained BERT model can grab different linguistic effects. Existing algorithms typically train the AS model with positive-negative pairs for a two-class classifier. A positive pair contains a question and a genuine answer, while a negative one includes a question and a fake answer. The output should be one for positive and zero for negative pairs. Typically, negative pairs are more than positive, leading to an imbalanced classification that drastically reduces system performance. To deal with it, we define classification as a sequential decision-making process in which the agent takes a sample at each step and classifies it. For each classification operation, the agent receives a reward, in which the prize of the majority class is less than the reward of the minority class. Ultimately, the agent finds the optimal value for the policy weights. We initialize the policy weights with the improved ABC algorithm. The initial value technique can prevent problems such as getting stuck in the local optimum. Although ABC serves well in most tasks, there is still a weakness in the ABC algorithm that disregards the fitness of related pairs of individuals in discovering a neighboring food source position. Therefore, this paper also proposes a mutual learning technique that modifies the produced candidate food source with the higher fitness between two individuals selected by a mutual learning factor. We tested our model on three datasets, LegalQA, TrecQA, and WikiQA, and the results show that RLAS-BIABC can be recognized as a state-of-the-art method.

## 1. Introduction

Today, the questions charged in numerous domains in cyberspace, such as Stack Overflow and GitHub, are progressing quotidianly. QA is one of the vital branches of natural language processing (NLP) that can have the ability to answer questions automatically. QA can be made in two ways: Several methods focus on generating answers that usually employ developing networks like generative adversarial network (GAN) to create answers [1]. Nonetheless, they cannot guarantee accurate meaning and grammar.

Another category of methods uses AS, one of the essential subtasks of QA which is also applied in other fields such as machine comprehension [2]. Over the last few years, the problem has been gaining an increasing amount of attention [3, 4]. A question $q$ and a set of candidate answers $A = \{a_1, a_2, a_3, \ldots, a_N\}$ are given, and the goal is to attain $a_i \in A$ as the best answer to question $q$. Questions and answers can have various lengths, and multiple answers may be the true answer to a question.

From the literature, there are numerous methods for AS based on traditional and deep learning methods [5]. The traditional approaches rely more on search engine [6], information retrieval [7, 8], handcrafted rules [9], or machine learning models [10, 11]. Information retrieval-based models work based on the keywords without using any semantic data, which makes it challenging to obtain the correct answers [12]. Handcrafted rule-based techniques cannot enfold all patterns, and their performance is delimited [13, 14]. In machine learning-based methods, features are manually made, so their quality laboriously depends on feature extraction [15, 16]. Some criteria and classifiers, including edit distance and support vector machine, consider the matching associations between AS pairs [11]. Typically, traditional methods suffer from two major weaknesses. First, they mostly do not use semantic information in keywords, features, or rules, causing them not to consider all-side relationships between QA pairs. Second, feature extraction and handmade rules are not flexible, leading to inferior generalization capability. After the appearance of deep learning, many problems in many domains [17–23], including AS, have been overshadowed by it. Deep learning-based methods for AS usually employ a convolutional neural network (CNN) [24] or LSTM to grab semantic features on various levels. The main task is to estimate the semantic similarity between a question-answer pair, which can be regarded as a text similarity calculation or classification work. A CNN is employed to model the hierarchical structures of sentences and evaluate their matching amount [25]. At the same time, an LSTM is considered to generate the embeddings of questions and answers while keeping sequential dependency information. Although deep models can only achieve limited improvement, they face some difficulties. They forge the embedding representation of the question-answer pair with one neural network design. This results in paying attention to one-side features and ignoring the other complex semantic features among question-answer couples. After that, models that try to comprehend languages were developed [26]. These models realize language syntactic and semantic rules in different methods, including next word and sentence prediction and masked word prediction [27]. They recognize a language and can make new texts with correct syntax and semantic rules. The BERT model [27] is one of the latest language models, being superior to all other developed language models. This model has grabbed advantage of the statement offered in transformers [28], which is currently widely employed in NLP tasks [29].

The success of deep models mainly relies on architecture, training algorithms, and selection of features employed in training. All these make the design of deep networks a complex optimization problem [30]. In many methods, the topology and transfer functions are set, and the space of possible networks is spanned by all potential values of the weights and biases [31]. In [32, 33] and [34], ant colony optimization [35], tabu search [36], simulated annealing [37], and genetic algorithm [38] were utilized for the training of neural networks with fixed topology. The neural network learning optimization process discovers the weight configuration associated with the lowest output error. Nevertheless, finding the optimal weight for deep models largely depends on weight initialization that has a more significant impact on neural network performance than network architecture and training examples [39]. AS methods, including in-depth ones, utilize gradient-based algorithms such as BP and Levenberg–Marquardt (LM) [40] for model weight optimization. While the BP algorithms converge in the first-order derivatives, the LM ones converge with second-order derivatives [41]. The main problem of BP and LM is the sensitivity to the initial weights, which leads to getting stuck in the local optimization [42]. To deal with this problem, global search approaches, having the power to evade local minima, are being employed to pretrain weights, such as population-based metaheuristic (PBMH) algorithms [43–45]. Among PBMH algorithms, ABC is one of the most powerful algorithms for optimization problems, which has two advantages over traditional algorithms: no need to calculate gradients and not getting caught up in local optimizations [46]. This algorithm is based on the intelligent behavior of bees, containing two general concepts: food sources and artificial bees. Artificial bees are looking for food sources with high nectar. The position of the food source shows a solution to the optimization problem, and the amount of nectar equals the quality of a solution. Although the food source position is a critical factor determining whether a bee selects a food source, some necessary information is still missing when bees produce a neighboring food source.

One of the other main problems in AS is imbalanced classes, since the member number of positive class, including the question and the corresponding answer, is much smaller than that of negative class, including the question and the non-corresponding answer, which reduces the performance of existing methods. Proposed methods with an imbalanced problem are generally divided into two groups: data-level methods and algorithmic-level methods. In data-level algorithms, training data is manipulated to balance class distribution by an oversampling minority class, undersampling majority class, or both. SMOTE [47] is an oversampling system that generates new examples by linear interpolation between adjacent minority samples. Near Miss [48] is an undersampling method that deals with an imbalanced problem by accidentally removing samples from a larger class. This algorithm eliminates the data of the larger class when viewing two data points belonging to two various classes that are close in terms of distribution. Oversampling algorithms can increase the possibility of overfitting, and undersampling algorithms lose valuable information in the majority class. In algorithmic-level methods, the importance of the minority class rises with techniques such as cost-sensitive learning, ensemble learning, and decision threshold adjustment. In the cost-sensitive learning methods, different costs are allocated to the wrong classification of each class in the loss function, which is more for the minority class. Ensemble learning-based solutions train multiple subclassifications and adopt voting to get better results. Threshold adjustment techniques train the classifier in the imbalanced dataset and change the decision threshold during the test. Deep learning-based methods have also been suggested to classify imbalanced data. The paper [49] introduced a loss function for deep models that equally receives classification errors from the majority and minority classes. Another study in [50] learns the discriminative features of imbalanced data while maintaining intercluster and interclass margins. The authors in [51] presented a method based on the bootstrapping algorithm that balances training data of convolutional network per mini-batch. An algorithm is proposed by [52] for optimizing network weights and class-sensitive costs. In [53], the authors extracted complex samples in the minority class and improved their algorithm by batchwise optimization with Class Rectification Loss function [54].

In the last few years, deep reinforcement learning has been successfully used in computer games, robots' control, recommendation systems [55–57], etc. For classification problems, deep reinforcement learning has helped eradicate noisy data and learn better features, which significantly improved classification performance. Nonetheless, little research has been accomplished on applying deep reinforcement learning to imbalanced classification. Deep reinforcement learning is ideally appropriate for imbalanced classification as its learning mechanism, and specific reward function is comfortable paying more attention to minority class by giving higher rewards or penalties.

This paper presents an attention mechanism-based LSTM model for AS, called RLAS-BIABC, established on the BERT word embedding, reinforcement learning, and an improved ABC algorithm. The main body of the RLAS-BIABC model consists of two attention-mechanism-based bidirectional LSTM (BLSTM)networks and a feedforward network to calculate the similarity of the question-answer pair. The model aims to learn both positive and negative pairs. The positive pair is related to the question and real answer, while the negative one considers each question with the other answers. We use BERT as word embedding to learn the semantic similarity between sentences without pre-engineered features. What is more, we introduce an improved ABC algorithm for RLAS-BIABC, whose task is to find weight initialization in all LSTMs, the attention mechanism, and feedforward network to begin the BP algorithm. In this regard, we modify the ABC algorithm by applying mutual learning between two selected position parameters to choose the candidate food source with higher fitness. In addition, in the BP step, our proposed method employs reinforcement learning to handle imbalanced classification in the proposed method. In this respect, we define the AS problem as a guessing game divided into a sequential decision-making process. At each step, the agent takes an environmental state represented by a training instance and then executes a two-class classification operation under the guidance of a policy. If the classifier accomplishes the operation well, it will take a positive reward; otherwise, it will take a negative reward. The minority class is more rewarded than the majority one. The agent's goal is to get as many cumulative rewards as possible during the sequential decision-making process, that is, to classify the samples as accurately as possible. We assess the RLAS-BIABC model on three standard datasets, TrecQA, LegalQA, and WikiQA, and show RLAS-BIABC to be superior to other methods that use random weighting.

The main contributions of the article are as follows: (1) We consider the BERT word embedding, which is the last developed model for many languages. (2) Instead of using the random weight system for the model weights, we define an encoding strategy and compute an initial value using an improved ABC algorithm. (3) We consider the AS problem a sequential decision-making process and propose a deep reinforcement learning framework for imbalanced classification. (4) We study the performance of the proposed model through experiments and compare it with the other methods that use the random weight for initialization and are faced with the imbalanced classification problem.

The rest of this article is organized as follows: Section 2 presents a short review of related works. Section 3 introduces the ABC algorithm. Section 4 describes the framwork of the proposed model. Section 5 exhibits evaluation metrics, datasets, andresults. Section 6 provides a conclusion and future works.

## 2. Related Work

Until now, many approaches to the QA problem have been proposed. This section provides an overview of the methods based on machine learning and deep learning.

The first proposed approaches were based on feature engineering. In these methods, the relationship between question and answer is measured by repeating common words, where bag-of-words and bag-of-grams [58] are commonly applied for this purpose. These methods are not logical because they do not respect semantic and linguistic features in sentences. Subsequently, however, some studies have utilized language resources such as WordNet [59] to resolve the semantic problem but failed to remove linguistic limitations. Some researchers considered sentences' syntactic and semantic structure [60]. Some authors employed the dependency tree and the tree edit distance algorithm [15, 61]. The research [62] confirmed that tools such as WordNet and NER [63] could play an influential role in selecting semantic features. The article [64] provided an effective solution for automated feature selection. These methods were one of the first attempts to eliminate feature engineering.

Later, with the advent of deep learning, many methods used deep models as an automatic feature engineering tool. Recently, in-depth learning has covered a wide range of applications of NLP[18]. Moreover, recurrent neural network (RNN) and CNN are applied as two strong arms of deep learning in feature extraction [20, 21]. The behavior of deep learning methods with question-answer pairs is divided into two categories. In the first category, question and answer are two distinct elements, and deep networks reach their representation vectors separately. Typically, various criteria are adopted to measure the similarity between them. The authors in [65] offered a compare-aggregate system that applies many metrics for similarity measuring. The study [66] utilized the ELMo language model [26] to overcome question and answer work. The results reveal the superiority of language models. In the second category, question and answer are assumed to be a single sentence. In [67], a CNN-based approach is presented to score question-answer pairs in a pointwise manner. Another technique in [68] applies the BLSTM network for question answering. Primarily, the embedding of question and answer words is learned and then entered into a BLSTM network, and later the embedding of each sentence is estimated based on the average of its words. Lastly, the answer-question connection is fed to a feedforward network. Siamese network [69] is an essential branch of in-depth learning that has been applied in all fields, especially QA. The network provides two separate representation vectors for question and answer. In [70], the first deep learning task is presented for the AS task. In this study, the most relevant answer to the question is extracted using a CNN and logistic regression. The research [71] implemented the idea presented in [70]. The authors tried to make different models using hidden layers, convolution operations, and activation functions to improve the results. Another work in [72] mixes various models to produce representation vectors for every sentence. In [73], the authors convert each point model into a pair model. Their idea was that pair models could further enhance model performance. The pair model was also applied to the model in [72]. The study [74] is a preprocessing operation. In this research, named entities are replaced with a unique token that facilitates selecting candidate answers. The impressive effectiveness of this technique was confirmed by applying it to the model presented in [73]. Meanwhile, the authors in [75] claimed that not all the named entities could be replaced with one token, so they considered a token for each named entity. It was later found that using the attention mechanism could produce more valuable models. Unlike the Siamese-based technique, the attention mechanism uses context-sensitive interactions [76] between question and answer. The attention mechanism was first proposed for machine translation but was later employed in other applications such as question answering [77, 78]. The approach in [79] considered the attention mechanism and RNNs to succeed in the answer-selection task. It was based on the attention mechanism proposed in [80]. In [81], the authors employed a method based on inter-weighted alignment networks to determine the similarity between a question-answer pair. The article [82] suggested a scheme based on a bidirectional alignment mechanism and stacked RNNs. In the first works, the attention mechanism was performed only on RNN, but later [83] pointed out that combining a CNN and attention mechanism could be more efficient.

## 3. Background

*3.1. Long Short-Term Memory (LSTM).* In a nutshell, RNNs [84] are designed to model sequential inputs. In these networks, a data sequence is mapped to a series of hidden states. The output is then generated using the following equations:

$$h_t = \theta(W_h h_{t-1} + U_h x_t + b_h). \quad (1)$$

$$y_t = \tau(W_y h_t + b_y), \quad (2)$$

where $W_h$ and $U_h$ are weight matrices and $b$ means bias. $\theta$ and $\tau$ represent the activation functions such as ReLU and Tanh. $x_t \in \mathbb{R}^d$ is the input with dimension $d$, and $h_t \in \mathbb{R}^h$ equals the hidden layer with size $h$ at time $t$.

RNNs have proven to be successful in many areas of NLP, such as text generation [85] and text summarization [86]. However, later, it became clear that as the length of the input of these networks increases, they suffer from problems such as gradient explosion and vanishing [87]. The LSTM network proposed by Hochreiter and Schmidhuber [88] can prevent the mentioned problems. This is because memory units can effectively handle long dependencies. In particular, LSTM consists of several control gates and one memory unit. Let $x_t$, $h_t$, and $c_t$ represent input, hidden state, and memory cell at time $t$, respectively. Given a sequence of inputs $(x_1, x_2, \ldots, x_T)$, LSTM should calculate a sequence of hidden units $(h_1, h_2, \ldots, h_T)$ and memory cells $(c_1, c_2, \ldots, c_T)$ as output. In terms of formula, the specified process can be defined as follows [89]:

$$\begin{aligned}
i_t &= \sigma(W_i x_t + U_i h_{t-1} + b_i), \\
f_t &= \sigma(W_f x_t + U_f h_{t-1} + b_f), \\
c_t &= f_t c_{t-1} + i_t \tanh(W_j x_t + U_j h_{t-1} + b_j), \\
o_t &= \sigma(W_o x_t + U_o h_{t-1} + b_o). \\
o_t &= \sigma(W_o x_t + U_o h_{t-1} + b_o), \\
h_t &= o_t \tanh(c_t),
\end{aligned} \quad (3)$$

where $W$ and $b$ are network parameters. $i$, $f$, and $o$ display input gate, forget gate, and output gate, respectively. $\sigma$ stands for sigmoid function.

Although many problems can be solved under the umbrella of LSTM networks [18, 19, 90], experiments show that BLSTM can be more effective than LSTM. A BLSTM network [91] is an extended LSTM net that processes input from start to end and vice versa. This process generates two hidden vectors, $\overrightarrow{h}_t$ and $\overleftarrow{h}_t$, for a specific input at the moment of $t$. Thus, the connected vectors, namely $[\overrightarrow{h}_t, \overleftarrow{h}_t]$, form the final hidden vector.

The information extracted by the units in the LSTM network is equally important in making the final decision, which reduces system performance. To illustrate the point, consider the sentence "Despite being from Uttar Pradesh, as she was brought up in Bengal, she is convenient in Bengali." In this sentence, words like "Bengali" and "Bengal" should be given more attention, while this is not the case in an original LSTM network. To overcome this problem, the attention mechanism has been considered. In an attention mechanism system, the importance of each hidden layer with a coefficient in the interval [0, 1] is involved in the construction of the final vector. Formally, the hidden unit vector for a particular input of length $T$ is calculated by considering the coefficient $\alpha_t$ for each hidden vector $h_t$ as follows:

$$h = \sum_{t=1}^{T} \alpha_t h_t. \qquad (4)$$

### 3.2. Artificial Bee Colony (ABC) Algorithm.

The ABC algorithm is a technique inspired by the intelligent behaviors of bees in nature. Two general concepts form the main body of the algorithm ABC: food sources and artificial bees. Artificial bees are looking for food sources with high nectar. The position of the food source indicates a solution to the optimization problem, and the amount of nectar corresponds to the quality of a solution. ABC involves three different groups of bees: employed, onlooker, and scout. Employed bees search for food sources with higher nectar in the vicinity of other food sources around them and share their information with onlooker bees in the dance area. The numbers of employed and onlooker bees are the same, and each is equal to half of the colony. Each employed bee exists in a hive, so the number of employed bees equals the total hives. Like employed bees, onlooker bees search for the best food sources in their neighborhood. Employed bees whose food resources do not improve after a few steps are converted to scouts, and a new search begins. The optimization process of ABC is summarized as follows:

*Initialization Stage.* Food sources as bee locations in the search space are initialized as follows:

$$x_i^j = x_{\min}^j + rand(0.1)(x_{\max}^j - x_{\min}^j). \qquad (5)$$

where $i$ refers to the $i$-th solution that takes the integer value in the interval $[1, BN]$, where $BN$ is the total number of solutions. Each solution consists of $D$ elements, where $D$ shows the number of weights to be optimized. $x_{\min}^j$ and $x_{\max}^j$ are the lowest and highest value in the solution $i$, respectively.

*Employed Bee Stage.* After initialization, the employed bees identify new sources in the neighborhood of existing food ones. Now they calculate the quality of the designated food sources. If their quality is better, they erase the information of previous sources from memory, replacing it with that of new sources. Otherwise, the data of earlier sources will remain unchanged. Formally, this step can be described by the following formula:

$$v_i^j = x_i^j + \varphi_i^j (x_i^j - x_k^j), \qquad (6)$$

where $k$ has an integer value in the interval $[1, BN]$, $\varphi_i^j$ is a random decimal value in $[-1, 1]$, and $v_i$ is a new food source derived from the change of an element $x_i$.

*Onlooker Bee Stage.* At this phase, the employed bees provide information to the onlooker bees. Onlooker bees calculate the value of the information and select the new solution based on the probability value. As in the previous step, if the new solution has more nectar, the previous position information will be replaced with the new solution. The possibility of choosing a new solution can be formulated as follows:

$$p_i = \frac{\text{fit}(x_i)}{\sum_{n=1}^{BN} \text{fit}(x_n)}, \qquad (7)$$

where $\text{fit}(xi)$ is the fitness value for the $i$-th solution. According to (7), the higher the $\text{fit}(xi)$ is, the more likely the observer bee will accept this solution. The onlooker bee goes to it if the selection is performed, and a new solution is generated according to (6).

*Scout Bee Stage.* In the last step, scout bees are employed to escape the local optimum. More specifically, any solution that fails to improve the process after some cycles becomes a scout bee, and the food source is dropped. Therefore, a new food source replaces the old one according to (6).

The four steps mentioned above are performed up to several times to meet the termination criteria. The complete ABC algorithm is given in Algorithm 1.

## 4. The Framework of RLAS-BIABC

The proposed algorithm considers two critical options for classification. In the first step, we formulate a vector that includes all the learnable weights in our model, and we optimize it utilizing the ABC algorithm. Then, we apply the BP algorithm in the rest of the path. Besides, another problem that most classifiers suffer from, including ours, is imbalanced data. To take this aspect into account, we employ the opinions of reinforcement learning. We present these two ideas in two separate sections.

The general architecture of the proposed model is shown in Figure 1. Consider a question $Q$ containing a sequence of $n$ words, where $Q = (q_1, q_2, \ldots, q_n)$, with the answer $A$, where $A = (a_1, a_2, \ldots, a_m)$ including $m$ words. Let $a_i, q_j \in \mathbb{R}^D$ show the $D$-dimensional visual presentations of a word. Two LSTMs are provided for each question and answer. Two pairs of positive and negative data are used to learn the model. In the positive pair $(Q, A)$, $A$ is the correct answer to question $Q$; the output of the model should go to one. Meanwhile, in the negative pair $(Q, A')$, where $A'$ is the fake answer to the question, the network should move to zero for this pair. The embedding calculated by LSTMs for question and answer is expressed as follows:

$$q = \sum_{i=1}^{n} \alpha_i h_{q_i},$$
$$a = \sum_{i=1}^{m} \beta_i h_{a_i}, \qquad (8)$$

where $h_{q_i} = [\overleftarrow{x}_i, \overrightarrow{x}_i]$ and $h_{a_i} = [\overleftarrow{y}_i, \overrightarrow{y}_i]$ are the output of $i$-th BLSTM related to the question and answer, respectively. $\alpha_i$ and $\beta_i$ are the attention weights of each unit that are computed as follows:

```
     Input: D: dimensions of the solution, BN: population size, limit: number of cycles, MaxItr: maximum number of iterations
(1)  Initialize the population of solutions X = [x_1, x_2, ..., x_BN] using (5)
(2)  Itr = 1.
(3)  while ≤ MaxItr do
(4)    //Employed Bee Phase
(5)    for i = 1 to BN do
(6)      Produce new solution x_new using (6)
(7)      Calculate the fitness f_new for x_new
(8)      Replace x_new with x_i if better
(9)    end for
(10)   Calculate the probability p for every solution in X using (7)
(11)   //Onlooker Bee Phase
(12)   for i = 1 to BN do
(13)     if rand (0, 1) < p_i then
(14)       Produce new solution x_new using (6)
(15)       Calculate the fitness f_new for x_new
(16)       Replace x_new with x_i if better
(17)     end if
(18)   end for
(19)   //Scout Bee Phase
(20)   If an abandoned solution is found, replace it with the solution produced by (6)
(21)   Put the best solution ever in x_best
(22)   Itr = Itr + 1.
(23) end while
(24) return x_best
```

ALGORITHM 1: Pseudocode of the ABC algorithm.

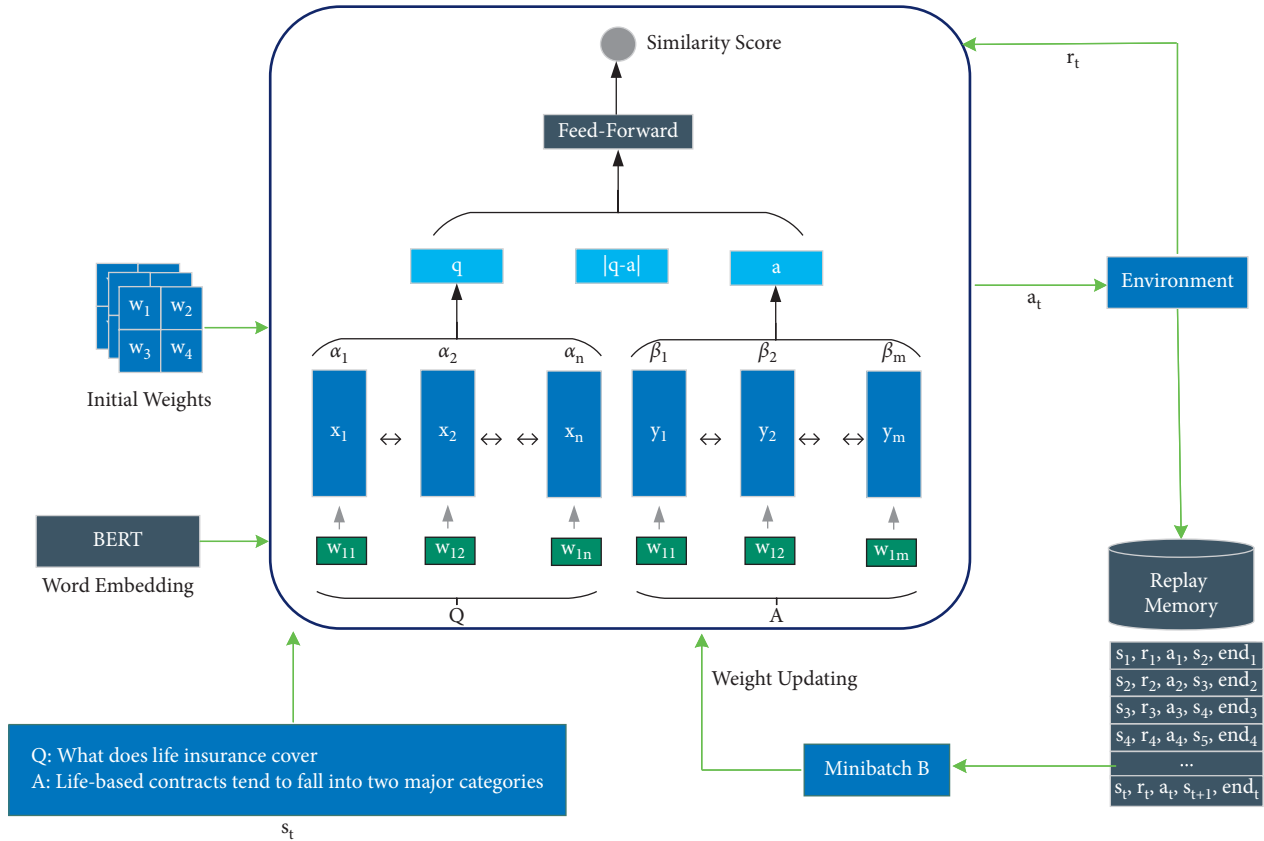

FIGURE 1: The proposed LSTM-similarity model.

$$\alpha_i = \frac{e^{u_i}}{\sum_{i=1}^{n} e^{u_i}},$$

$$\beta_i = \frac{e^{v_i}}{\sum_{i=1}^{m} e^{v_i}}, \quad (9)$$

$$u_i = \tanh\left(W_u\left[\overleftarrow{x}_i, \overrightarrow{x}_i\right] + b_u\right),$$

$$v_i = \tanh\left(W_v\left[\overleftarrow{y}_i, \overrightarrow{y}_i\right] + b_v\right),$$

where $W_u, W_v, b_u, b_v$ represent the parameters of the attention mechanism. After determining the efficient representation of question and answer by the attention mechanism, we form a vector consisting of the connected $q$, $a$, and $|q - a|$ according to Figure 1 and enter it into a feedforward network. It has been experimentally confirmed that the difference between two representation vectors can act in a successful decision [92].

### 4.1. BERT-Based Word Embedding.

Word embedding serves as a function of mapping words to semantic vectors for use in deep learning algorithms. Word embedding is a reliable way to extract significant representations of words established in their context. Much research has been conducted to find the best meaningful word representations on neural network models such as Skip-gram [93], GloVe [94], and FastText [95]. Over the last few years, the pretrained language model (PLM), which is a black box with prior knowledge of the natural language and is fine-tuned in NLP works, has been much applied.

PLM models generally use unlabeled data to learn model parameters [96]. The paper considers the BERT model [27], one of the latest techniques in the PLM trends. BERT is a bidirectional language model trained on big datasets such as Wikipedia to generate contextual representations. In addition, it is commonly fine-tuned from a neural network dense layer for different classification duties. The fine-tuning functionality includes the contextual or the problem-specific meaning with the pretrained generic meaning and trains it for a classification task.

Figure 2 indicates the architecture of a BERT model. BERT uses a bidirectional transformer, in which its representations are jointly conditioned on both the left and right context in layers [97], which differentiates it from the other models, including Word2Vec and GloVe, that build an embedding in one direction to dismiss its contextual differences.

### 4.2. Pretraining Stage.

Weight initialization is an essential point in designing a neural network, the nonobservance of which leads to misleading the model. The proposed structure has two LSTM networks, two attention mechanisms, and one feedforward neural network, each of which has its weights that must be trained. The paper uses an improved ABC algorithm for pretraining weights.

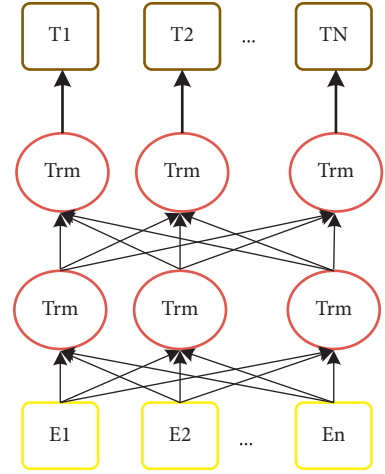

Figure 2: Architecture of the BERT model.

*4.2.1. Mutual Learning-Based ABC.* In the standard ABC algorithm, artificial bees randomly select a food source position and change it to create a new position. If the fitness value of the new solution is better, it will replace the current solution. Otherwise, no change will be applied. In other words, in a $D$-dimensional optimization problem, one dimension is randomly selected, its value is changed, and the better outcome is selected in each iteration. Based on (6), the newly generated solution $v_i^j$ depends on only two parameters, $x_i^j$ and $x_k^j$, making the food source $v_i^j$ uncontrollable, sometimes larger and sometimes smaller than the current food source. In the ABC algorithm, a food source with a higher fitness value is required. To always produce a food source a higher value, we consider the fitness information acquired by mutual learning between current and neighboring food sources.

$$v_i^j = \begin{cases} x_i^j + \varphi_i^j\left(x_k^j - x_i^j\right), & \text{Fit}_i < \text{Fit}_k \\ x_k^j + \varphi_i^j\left(x_i^j - x_k^j\right), & \text{Fit}_i \geq \text{Fit}_k \end{cases}, \quad (10)$$

where $\text{Fit}_i$ and $\text{Fit}_k$ indicate the fitness value of the current food source and the neighboring food source, respectively. $\varphi_i^j$ shows a uniform random number in the interval $[0, F]$, in which $F$ is a nonnegative constant named the mutual learning factor. As we can see, the value $v_i^j$ depends on their position and their value of fitness. By comparing the current and neighboring food sources, the fitness values of new solutions move to better sources. That is, if the current food source has higher suitability, the candidate solution will move toward a better solution; otherwise, it will tend to move toward the neighboring source. This learning strategy allows making a better candidate solution. The parameter $F$ plays an essential role in balancing the perturbation between related food positions. In addition, $F$ must be a nonnegative positive number to ensure it goes to a better solution. As $F$ increases from zero to a particular value, the perturbation on the corresponding position decreases, meaning that the fitness value of the new food source is close to the higher fitness. A large value of $F$ weakens the power of exploitation and exploration.

*4.2.2. Encoding Strategy.* Encoding means the weights are arranged in a vector, which is considered the bees' position in ABC. Choosing the right layout is a challenging task; however, we tried to design the best encoding strategy possible after several experiments. Figure 3 denotes an example of the encoding for two LSTMs, two attention mechanisms, and a two-layer feedforward network. Note that all weight matrices are stored in rows.

*4.2.3. Fitness Function.* The purpose of the fitness function is to measure the efficiency of a solution. The paper employs the following function as a competency function:

$$\text{fitness} = \frac{1}{1 + \sum_{i=0}^{T}(y_i - \tilde{y}_i)^2}, \quad (11)$$

where $T$ is the total number of training samples and $y_i$ and $\tilde{y}_i$ are the target and predicted labels for the $i$-th data, respectively.

*4.3. Classification.* Reinforcement learning (RL) [98] is a subfield of machine learning that solves a problem by making successive decisions [99, 100]. Recently, reinforcement learning has achieved excellent results in classification because it can learn valuable features or select high-level samples from noise data. In [101], the classification problem was defined as a sequential decision-making process that used several factors to learn the optimal policy. However, complex simulations between agents and environments have somewhat increased the time complexity. Another work in [102] submitted a solution for learning a relationship in text noise data. For this purpose, the proposed model is divided into two parts: instance selector and relational classifier. The instance selector is designed to extract quality sentences from noise data with the agent help. At the same time, the relational classifier learns better performance from selected clean data and gives delayed reward feedback to the instance selector. Finally, the model results in a better classification and quality dataset. The authors in [103–106] considered deep reinforcement learning to learn the helpful training data features. Generally, they improved the valuable features of the classifier. The work in [107] used reinforcement learning to classify time series data in which the reward function and the Markov model are designed. So far, little research has been done on the classification of unbalanced data, especially the processing of natural languages using reinforcement learning. In [108], an ensemble pruning method that picks the best sub-classifiers under the reinforcing learning umbrella was developed. This method was effective for small data because it was practically impossible to choose classifiers with many subcategories.

This section describes how to apply reinforcement learning to prevent imbalanced classification. Overall, the agent receives a sample at each step and classifies it. After that, the environment gives immediate and next rewards to the agent. A positive reward is assigned to the agent by the environment when it categorizes the sample correctly. Otherwise, it receives a negative reward. Finally, the agent learns the optimal behavior by maximizing the aggregate rewards and then can classify the samples as accurately as possible.

Let $D = \{(x_1, l_1), (x_2, l_2), \ldots, (x_T, l_T)\}$ be training data, where $x_i = (q_i, a_i)$ is the $i$-th sample so that $q_i$ and $a_i$ are the $i$-th question and answer that enter the model, respectively. $l_i \in \{0, 1\}$ shows the target of the $i$-th example. We consider the following conditions for an agent.

*4.3.1. Policy $\pi_\theta$.* The policy $\pi_\theta$ is a mapping function $\pi: S \longrightarrow A$ where $\pi_\theta(s_t)$ denotes the action $a_t$ performed by an agent in state $s_t$. In our work, the proposed classification with the set weight $\theta$ is recognized as policy $\pi_\theta$.

*4.3.2. State $s_t$.* Each example of the training dataset is described as a state. The agent takes the first data $x_1$ as the initial state $s_1$ at the start of the training. State $s_t$ at each time step $t$ corresponds to $x_t$ in the training dataset. The order of the samples in each iteration is different for the agent.

*4.3.3. Action $a_t$.* The action performed by the agent is to predict the category label. Hence, the agent's performance is related to the training dataset label. The recommended model is a binary classifier, $a_t \in \{0, 1\}$, where zero and one show the minority and majority classes, respectively. In this context, the relevant question and answer are one, and the irrelevant question and answer are zero.

*4.3.4. Reward $r_t$.* The agent receives a positive score if the sample is classified correctly and a negative score otherwise. Since minority class instances are more critical because of their small number, the algorithm should consider the size of the score for the minority class more. The reward function is described as follows:

$$r(s_t, a_t, l_t) = \begin{cases} +1, & a_t = l_t \text{ and } s_t \in D_P \\ -1, & a_t \neq l_t \text{ and } s_t \in D_P \\ \lambda, & a_t = l_t \text{ and } s_t \in D_N \\ -\lambda, & a_t \neq l_t \text{ and } s_t \in D_N \end{cases}, \quad (12)$$

where $\lambda \in [0, 1]$, and $D_P$ and $D_N$ are related to the minority and majority classes, respectively. $l_t$ is the label of the sample $x_t$. The bonus amount is considered the cost of predicting the label. According to this relation, when $\lambda < 1$, the amount of the cost of the minority class is more. If the distribution of all classes is balanced, $\lambda = 1$, then the prediction cost of all classes is the same. We will examine the different values of $\lambda$ in our experiments.

*4.3.5. Terminal E.* The episode is a transition trajectory from the initial state to the terminal state $\{(s_1, a_1, l_1), (s_2, a_2, l_2), \ldots, (s_t, a_t, l_t)\}$. An episode finishes when all instances in the training data are classified or when the agent misclassifies the instance from the minority class.

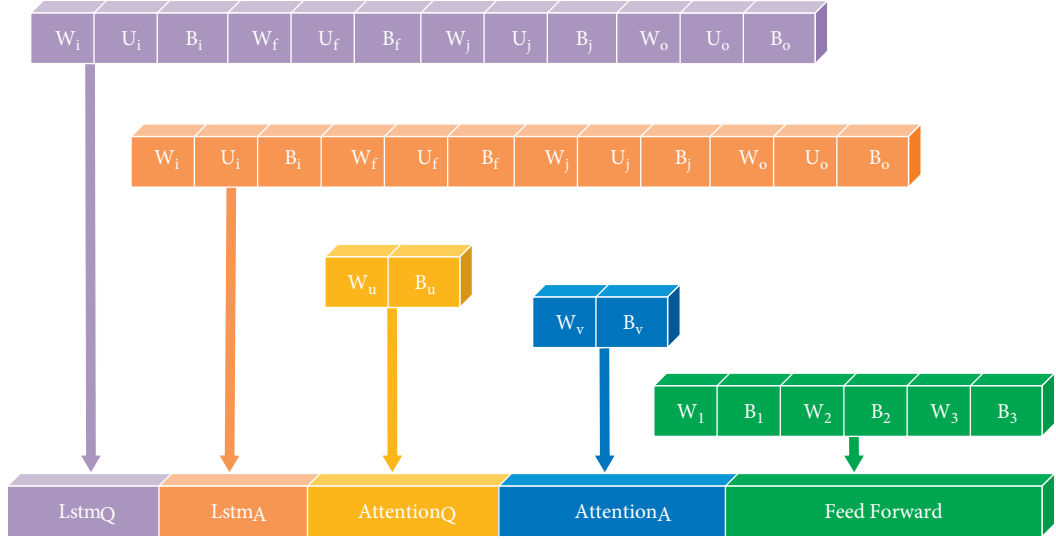

Figure 3: Placement of weights in a vector.

*4.3.6. Transition Probability P.* The model transition probability, i.e., $p(s_{t+1}|s_t, a_t)$, is deterministic. The agent transfers from state $s_t$ to state $s_{t+1}$ according to the order of instances in the dataset.

In the proposed model, the $\pi$ policy takes the input data and calculates its label probability:

$$\pi(a \mid s) = P \cdot (a_t = a \mid s_t = s). \quad (13)$$

The agent aims to identify the data input sample as accurately as possible. The best performance is attributed to the agent when it can maximize its cumulative rewards as follows:

$$g_t = \sum_{k=0}^{\infty} \gamma^k r_{t+k}. \quad (14)$$

Equation (14) is called the return function, the total accumulated return from time $t$ with the discount factor $\gamma \in (0, 1]$ until the time when the agent moves in the search space. The action value $Q$ in RL expresses the expected return for action $a$ in state $s$, which can be defined as follows:

$$Q^\pi(s, a) = E_\pi[g_t \mid s_t = s, a_t = a]. \quad (15)$$

Equation (15) can be extended according to the Bellman Equation [109]:

$$Q^\pi(s, a) = E_\pi[r_t + \gamma Q^\pi(s_{t+1}, a_{t+1}) \mid s_t = a, a_t = a]. \quad (16)$$

By maximizing function $Q$ under policy $\pi$, we can maximize cumulative rewards, namely $Q^*$. The optimal policy $\pi^*$ obtained under function $Q^*$, which is a policy that performs best for our model, is as follows:

$$\pi^*(a \mid s) = \begin{cases} 1, & a = \mathrm{argmax}_a Q^*(s, a) \\ 0, & \text{else} \end{cases}. \quad (17)$$

By combining (16) and (17), function $Q^*$ is computed as follows:

$$Q^*(s, a) = E_\pi[r_t + \gamma \mathrm{max}_a Q^*(s_{t+1}, a_{t+1}) \mid s_t = a, a_t = a]. \quad (18)$$

For low dimensions, the values of the function $Q$ are collected in a table to obtain the optimal value according to the recorded values. However, the function $Q$ can no longer be solved when the dimensions of the problem are continuous. To solve this problem, a deep $Q$-learning algorithm was adopted to model the function $Q$ with a deep neural network. To that end, the tuple $(s, a, r, s')$ obtained from (18) is stored in replay memory $M$. The agent selects a mini-batch $B$ of transitions from $M$ randomly and executes the dissent gradient algorithm on the deep $Q$ network according to the following loss function:

$$L(\theta_k) = \sum_{(s,a,r,s') \in B} (y - Q(s, a; \theta_k))^2, \quad (19)$$

where $y$ is the prediction of the function $Q$, which is formulated as follows:

$$y = \begin{cases} r, & \text{end} = \text{True} \\ r + \gamma \mathrm{max}_{a'} Q(s', a'; \theta_{k-1}), & \text{end} = \text{True} \end{cases}, \quad (20)$$

where $s'$ indicates the next state $s$, and $a'$ is the action executed in state $s'$.

*4.4. Overall Algorithm.* We design the simulation environment according to the contents defined above. The network architecture of the policy largely depends on the complexity and number of training examples. In this context, the input of the network depends on the structure of the training samples, and the output is equal to the number of classes of instance data. The general training algorithm of the model presented in Algorithm 2 is shown. First, the initial weights of the policy $\pi$ are initialized using the ABC algorithm, and then the agent continues the training process until the optimal policy is reached. The choice of action is made based on the greedy policy, and the selected action is evaluated by Algorithm 3. The algorithm is repeated $E$ times, where $E$ in this paper is considered 15,000. At each step, the policy network weights are stored.

## 5. Results

*5.1. Datasets.* A dataset with many negative pairs can be one of the best options to evaluate the performance of the proposed system. We run our experiments on three datasets, LegalQA, TrecQA, and WikiQA, which are widely considered by many researchers. All three datasets have more negative than positive pairs. The statistical information of all datasets is shown in Table 1:

(i) TrecQA [110] is derived from TREC track data. Yao et al. [10] made a complete version of the positive and negative pair set. Two training datasets, TRAIN and TRAIN-ALL, are available in this database. The correctness of the answers in TRAIN-ALL is checked automatically by matching pairs with regular expressions. All answers in the TRAIN, DEV, and TEST data were judged manually. We employ the TRAIN-ALL data to train our model.

(ii) LegalQA [111] is a Chinese dataset of legal consultative questions collected from a Chinese association. Users' online questions have been answered by licensed lawyers. LegalQA includes four fields: question subject, question body, answer, and label. The positive pair is provided as ground truth directly online.

(iii) WikiQA [112] is an open-domain QA dataset in which each question is linked to a Wikipedia page that is assumed to be the topic of the year. To eliminate answer sentence prejudice, all answers in the summary section of the page are considered candidate answers.

*5.2. Evaluation Metrics.* According to previous research, MAP and MRR are the most common criteria for evaluating answer-selection tasks [77]. MAP measures the ability to rank answers to return the corresponding answer. However, MRR is repeated if a high-scoring match is found:

(i) MAP (mean average precision) calculates the mean average precision on the ranking results as follows:

$$\text{MAP}(Q) = \frac{1}{|Q|} \sum_{i=1}^{|Q|} \frac{1}{n_i} \sum_{j=1}^{n_i} \text{precision}(R_{ij}), \quad (21)$$

where Q denotes the set of questions, $n_i$ is the number of answers to the $i$-th question, and $R_{ij}$ means the set of ranked results to question $j$ from the best result to the $j$-th answer.

(ii) MRR (mean reciprocal rank) evaluates the model suitability according to the position of the first correct answer, computed as follows:

$$\text{MRR}(Q) = \frac{1}{|Q|} \sum_{i=1}^{|Q|} \frac{1}{r_i}, \quad (22)$$

where $r_i$ indicates the position of the first matching answer for the $i$-th question.

*5.3. Baseline Methods.* We evaluate our RLAS-BIABC model with several state-of-the-art methods for answer selection. The following are the details of these methods:

KABLSTM [113] is a knowledge-aware method based on attentive BLSTM networks. This method uses knowledge graphs (KG) to learn the representation of questions and answers.

EATS [75] adopted an RNN network to measure the similarity between the QA pair. First, it replaces each named entity with a specific word. This system calculates sentence representation vectors by the attention mechanism. Finally, these vectors are entered into the feedforward network, and the similarity is calculated by the sigmoid function in the last layer.

AM-BLSTM [114] considered two LSTM networks for a question and answer separately. The resulting embeddings were combined and entered into an multilayer perceptron (MLP) network for classification. Moreover, traditional techniques, such as penalties for each class, have been employed to prevent imbalanced classification.

BERT-Base [115] introduced a search engine and transformer model method for selecting answers. This article adopts simple models such as Jaccard similarity and compare-aggregate to rank the answers to a question.

DRCN [116] offered an architecture based on a densely connected recurrent and co-attentive network in which hidden features are maintained at the top layer. Connection operations in this paper are performed using the attention mechanism to preserve information better. In addition, an autoencoder has been adopted to reduce the volume of information.

P-CNN [117] introduced a new approach using a positional CNN for text matching that considers positional information at the word, phrase, and sentence levels.

DARCNN [118] combined BLSTM, self-attention, cross-attention, and CNN to find the global and local features of the question and candidate answer, leading to better semantic modeling. Finally, it utilizes an MLP network to assign a score to a question-answer pair.

DASL [119] submitted a model with a Bayesian neural network (BNN) to effectively optimize the loss in the ranking learning process. Another study of this article is how to combine active learning and self-paced learning for model training.

KAAS [120] applied an interactive knowledge-enhanced attention network for AS that extracts rich features of question and answer knowledge at several levels. Additionally, an attention and self-attention network is considered to learn the semantic features of sentences.

```
        Input: D = {(x_1, l_1), (x_2, l_2), ..., (x_T, l_T)}: a training dataset of size T
  (1)   Initialize the weights of policy π using Algorithm 1
  (2)   Initialize environment ε
  (3)   Initialize replay memory M
  (4)   for episode e = 1 to E do
  (5)       Shuffle the dataset D
  (6)       s_1 = x_1
  (7)       for t = 1 to T do
  (8)           a_t = π_0(s_t)  //select an action based on ε-greedy
  (9)           [r_t, end_t] = Reward(x_t, a_t, l_t)
  (10)          s_{t+1} = x_{t+1}
  (11)          Save (s_t, a_t, r_1, s_{t+1}, end_t) to M
  (12)          Sample randomly a mini-batch of transitions (s_k, a_k, r_k, s_{k+1}) from M
  (13)          y_k = { r_k,                              end_k = True
                      { r_k + γ max_{a'} Q(s_{k+1}, a'; θ),  end_k = False
  (14)          Accumulate gradients w.r.t θ: dθ = dθ + ∂(y_k − Q(s, a; θ))^2/∂θ
  (15)          if end_t = True then
  (16)              break
  (17)          end if
  (18)      end for
  (19)  end for
```

ALGORITHM 2: Pseudocode for training RIAS-BIABC.

```
        Function Reward (x_t, a_t, l_t):
  (1)       end_t = False
  (2)       if x_t ∈ D_p then
  (3)           if a_t == l_t then
  (4)               r_t = 1
  (5)           else
  (6)               r_t = −1
  (7)               end_t = True
  (8)           end if
  (9)       else:
  (10)          if a_t == l_t then
  (11)              r_t = λ
  (12)          else:
  (13)              r_t = −λ
  (14)          end if
  (15)      end if
  (16)      return [r_t, end_t]
```

ALGORITHM 3: Pseudocode of reward function.

TABLE 1: Statistical information of LegalQA, TrecQA, and WikiQA datasets.

| Dataset (TRAIN/DEV/TEST) | # questions | # QA pairs | % correct |
|---|---|---|---|
| LegalQA | 10,526/1,593/3,035 | 100,590/11,965/26,913 | 21.8/24.4/22.9 |
| TrecQA | 1,229/82/100 | 53,417/1,148/1,517 | 12.0/19.3/18.7 |
| WikiQA | 873/126/243 | 20,360/1,130/2,352 | 12.0/12.4/12.5 |

"% correct" means the proportion of matched QA pairs.

5.4. Details of Implementation. In this work, Python and PyTorch have been utilized for the implementation. Jupyter has been used to implement project codes. Another library used in this study is NLTK. This library provides classes and methods for processing natural languages in Python. This library can perform a wide range of NLP operations. We use a two-layer BLSTM. Moreover, due to the connection of vectors in the two networks, we employ batch normalization before the data enters the feedforward neural network. Table 2 indicates the values of the other parameters.

Our project uses a 64-bit Windows operating system with 64 GB of RAM and GPU. The best model was obtained for the LegalQA, TrecQA, and WikiQA after 50, 60, and 100 epochs, respectively. The whole process of our training took 5, 20, and 60 hours for the three datasets.

5.5. *Experimental Results.* Due to heuristic algorithms working randomly, we repeated all the experiments 10 times. Quantitative results of the three datasets are given in Table 3. In addition to comparing the proposed method with the state-of-the-art algorithms, to evaluate the effectiveness of ABC and RL components on the model, we employ three techniques: AS + random weight, AS-BIABC, and RLAS. AS + random weight is a system applying only random weights for initial weighting. Models AS-BIABC and RLAS accept only ABC and RL, respectively. For the LegalQA dataset, the RLAS-BIABC model has beaten other models, including IKAAS, in the MAP and MRR criteria, so that our model has reduced the error by more than 40% and 24% in these two criteria. By comparing RLAS-BIABC with AS-BIABC and RLAS, we can see that it decreases the error rate by about 51%, indicating the importance of the initialization and RL approaches. For the TrecQA dataset, our algorithm obtained the highest MAP and MRR, followed by EAT algorithm. The error improving rate in this database is approximately 30.13% and 21.00% for MAP and MRR criteria, respectively. In the WikiQA dataset, RLAS-BIABC decreases the classification error by more than 32% and 42% compared to IKAAS and DRCN, respectively.

Next, we prove that the improved ABC is more powerful than others. To do this, we fix all pieces of our algorithm for a fair comparison, including the LSTM networks, the attention mechanisms, and the reinforcement learning, and only change the trainer. To reach this goal, we compare our offered trainer with six conventional algorithms, including GDM [121], GDA [122], GDMA [123], OSS [124], and BR [125], and eight metaheuristic algorithms, including GWO [126], BAT [127], DA [128], SSA [129], COA [130], HMS [131], WOA [132], and ABC [133]. In all metaheuristic methods, population size and function evaluations are 100 and 3,000, respectively. The rest of the parameters of the algorithms are shown in Table 4. The results of metaheuristic and conventional algorithms are collected in Table 5. RLAS-AM-BR and RLAS-BABC performed best for all datasets for conventional and metaheuristic algorithms. As we expected, the metaheuristic algorithms perform better than the conventional ones. Without exaggeration, the improved ABC has a more acceptable performance than all of them, so that compared to the best algorithm, i.e., the main version of ABC, it can diminish the error by approximately 16%.

5.5.1. *The Effect of the Reward Value of Majority Class.* The environment helps the agent achieve the goal by considering the reward function. This article considers two different rewards for the minority and majority classes. Minority class reward was set to +1/−1 while the majority class was set to +$\lambda$/−$\lambda$. To investigate the effect of the value of $\lambda$ on the proposed model, we test it with the values in the set

Table 2: The parameters of the model.

| Parameter | Value |
| --- | --- |
| Batch size | 128 |
| Embedding dim | 60 |
| Max sentence length | 80 |
| Activation function (LSTM and dense) | ReLU |
| Dense hidden layer | 8 |

{0.1, 0.2, 0.3, 0.4, 0.5, 0.6, 0.7, 0.8, 0.9, 1}. The results of this experiment for the three datasets are indicated in Figure 4. As we see, for the LegalQA dataset, when $\lambda$ has a value in the range [0, 0.4], we have an uptrend, while we have a downtrend for the values (0.4, 1]. Hence, we fixed the value of $\lambda$ for this dataset to 0.5. The best value of $\lambda$ for both TrecQA and WikiQA datasets is 0.5. Generally, as the dataset size increases, the number of negative pairs increases, so $\lambda$ tends to decrease. For $\lambda = 0$, the importance of the majority class is overlooked, and for most $\lambda = 1$, the importance of both classes is equal.

5.5.2. *Exploration on Loss Function.* Traditional techniques, including manipulating the loss function and data augmentation, can also deal with data imbalances. However, they largely depend on the issue at hand. In the meantime, the loss function has a more colorful role because it can make the minority class more prominent. To check the inefficiency of the loss functions on our model, we selected the five functions Weighted Cross-Entropy (WCE) [134], Balanced Cross-Entropy (BCE) [135], Focal Loss (FL) [136], Dice Loss (DL) [137], and Tversky Loss (TL) [138]. The WCE and BCE loss functions give weight to the positive and negative samples. The FL function is suitable for applications with imbalanced data. It downweights the contribution of uncomplicated examples and allows the model to focus more on learning complex samples [139]. The evaluation results of these loss functions for the three datasets are shown in Table 6. The results show that all the functions have about the same MRR and MAP in the three datasets. As expected, the FL function performs better than the others, so it is about 51.16% better than the algorithm with the usual loss function, i.e., the RLAS-BABC model.

5.5.3. *Case Study.* In this section, we intend to qualitatively evaluate the effectiveness of reinforcement learning in our model. For this purpose, we randomly select a sample from the TrecQA dataset. Given the question, "When were the Nobel Prize awards first given?" top answers are given in Table 7. The left column presents the model results without using reinforcement learning, and the right column shows the model results with reinforcement learning. Our results say that the model without reinforcement learning is more inclined to assign a higher score to negative responses. However, the model with reinforcement learning has assigned as many scores as possible to the answers to the question.

Table 3: The evaluation results of the proposed model and other models.

| Method | LegalQA | | TrecQA | | WikiQA | |
|---|---|---|---|---|---|---|
| | MAP | MRR | MAP | MRR | MAP | MRR |
| KABLSTM [113] | 0.751 | 0.790 | 0.792† | 0.844† | 0.732† | 0.749† |
| EATS [75] | 0.778 | 0.810 | 0.854† | 0.881† | 0.700† | 0.715† |
| AM-BLSTM [114] | 0.786 | 0.836 | 0.818 | 0.827 | 0.780 | 0.788 |
| BERT-Base [115] | 0.838 | 0.850 | 0.823 | 0.812 | 0.813† | 0.828† |
| DRCN [116] | 0.828 | 0.859 | 0.802 | 0.832 | 0.804† | 0.862† |
| P-CNN [117] | 0.715 | 0.729 | 0.680 | 0.698 | 0.734† | 0.737† |
| DARCNN [118] | 0.700 | 0.745 | 0.743 | 0.725 | 0.734† | 0.750† |
| DASL [119] | 0.804 | 0.816 | 0.824 | 0.831 | 0.768 | 0.795 |
| IKAAS [120] | 0.825† | 0.883† | 0.823† | 0.868† | 0.835 | 0.849 |
| AS + random weight | 0.758 ± 0.000 | 0.801 ± 0.001 | 0.796 ± 0.000 | 0.806 ± 0.002 | 0.771 ± 0.002 | 0.792 ± 0.009 |
| AS-BIABC | 0.788 ± 0.012 | 0.815 ± 0.008 | 0.802 ± 0.005 | 0.826 ± 0.002 | 0.803 ± 0.000 | 0.845 ± 0.025 |
| RLAS | 0.855 ± 0.102 | 0.872 ± 0.018 | 0.862 ± 0.014 | 0.883 ± 0.150 | 0.852 ± 0.025 | 0.876 ± 0.026 |
| RLAS-BIABC | 0.895 ± 0.020 | 0.912 ± 0.001 | 0.898 ± 0.015 | 0.906 ± 0.092 | 0.888 ± 0.036 | 0.891 ± 0.017 |

† indicates that the results are taken from the articles.

Table 4: Parameter setting of experiments.

| Algorithm | Parameter | Value |
|---|---|---|
| ABC | Limit | $n_e \times$ dimensionality of problem |
| | $n_o$ | 50% of the colony |
| | $n_e$ | 50% of the colony |
| | $n_s$ | 1 |
| GWO | No parameters | |
| BAT | Constant for loudness update | 0.4 |
| | Constant for an emission rate update | 0.6 |
| | Initial pulse emission rate | 0.002 |
| DA | Scaling factor | 0.3 |
| | Crossover probability | 0.7 |
| SSA | No parameters | |
| COA | Discovery rate of alien solutions | |
| HMS | Number of clusters | 5 |
| | Minimum mental processes | 2 |
| | Maximum mental processes | 5 |
| WOA | C | 1 |
| | B | 1 |

Table 5: The performance of other methods for initialization.

| Method | LegalQA | | TrecQA | | WikiQA | |
|---|---|---|---|---|---|---|
| | MAP | MRR | MAP | MRR | MAP | MRR |
| RLAS-BGDM | 0.796 ± 0.002 | 0.819 ± 0.026 | 0.824 ± 0.093 | 0.836 ± 0.026 | 0.810 ± 0.056 | 0.825 ± 0.136 |
| RLAS-BGDA | 0.783 ± 0.125 | 0.776 ± 0.095 | 0.769 ± 0.025 | 0.786 ± 0.269 | 0.745 ± 0.136 | 0.761 ± 0.002 |
| RLAS-BGDMA | 0.791 ± 0.005 | 0.772 ± 0.103 | 0.796 ± 0.126 | 0.812 ± 0.236 | 0.793 ± 0.026 | 0.793 ± 0.005 |
| RLAS-BOSS | 0.810 ± 0.136 | 0.814 ± 0.004 | 0.853 ± 0.023 | 0.863 ± 0.026 | 0.840 ± 0.027 | 0.855 ± 0.127 |
| RLAS-BBR | 0.842 ± 0.009 | 0.853 ± 0.000 | 0.860 ± 0.036 | 0.878 ± 0.120 | 0.852 ± 0.103 | 0.870 ± 0.035 |
| RLAS-BGWO | 0.771 ± 0.205 | 0.783 ± 0.018 | 0.755 ± 0.072 | 0.781 ± 0.126 | 0.755 ± 0.025 | 0.773 ± 0.026 |
| RLAS-BBAT | 0.862 ± 0.003 | 0.818 ± 0.019 | 0.876 ± 0.093 | 0.880 ± 0.239 | 0.852 ± 0.061 | 0.873 ± 0.082 |
| RLAS-BDA | 0.816 ± 0.072 | 0.829 ± 0.022 | 0.863 ± 0.002 | 0.883 ± 0.056 | 0.836 ± 0.082 | 0.862 ± 0.091 |
| RLAS-BSSA | 0.747 ± 0.029 | 0.769 ± 0.072 | 0.750 ± 0.042 | 0.763 ± 0.025 | 0.746 ± 0.041 | 0.755 ± 0.001 |
| RLAS-BCOA | 0.860 ± 0.085 | 0.889 ± 0.089 | 0.882 ± 0.063 | 0.897 ± 0.237 | 0.872 ± 0.093 | 0.862 ± 0.017 |
| RLAS-BHMS | 0.849 ± 0.002 | 0.880 ± 0.123 | 0.879 ± 0.090 | 0.893 ± 0.036 | 0.840 ± 0.100 | 0.870 ± 0.009 |
| RLAS-BGDM | 0.752 ± 0.012 | 0.753 ± 0.027 | 0.769 ± 0.058 | 0.789 ± 0.085 | 0.731 ± 0.000 | 0.760 ± 0.018 |
| RLAS-BABC | 0.875 ± 0.004 | 0.906 ± 0.021 | 0.888 ± 0.046 | 0.900 ± 0.082 | 0.878 ± 0.016 | 0.889 ± 0.023 |

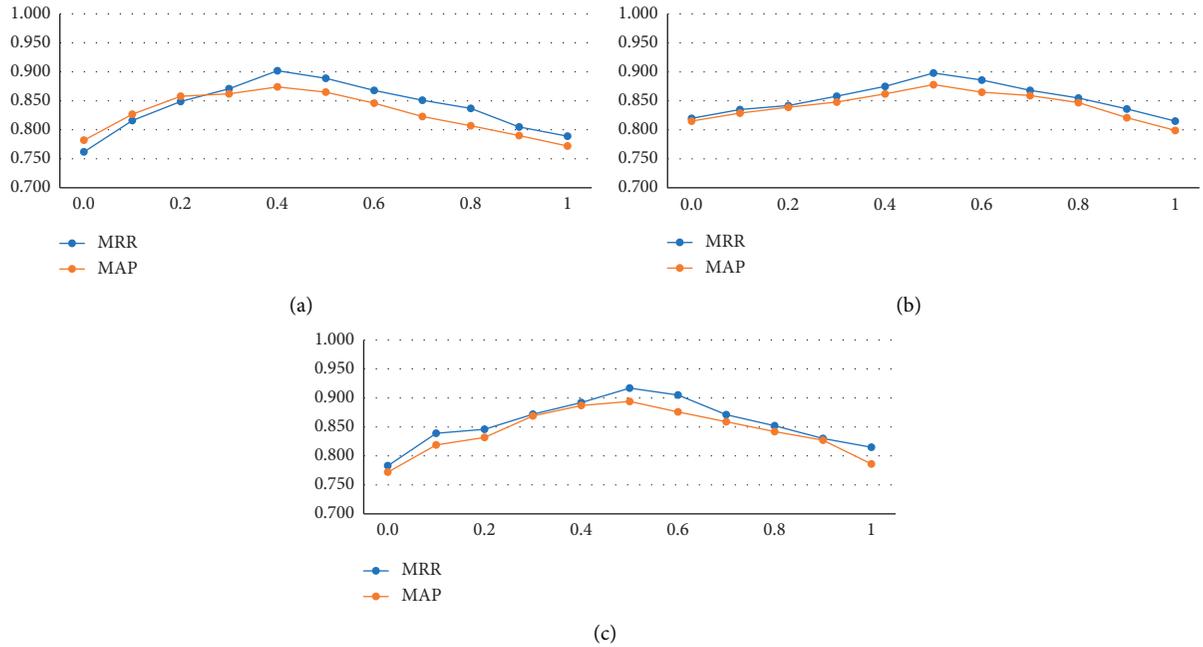

Figure 4: The process of changing the criteria by modifying the value of $\lambda$ for the three datasets: (a) LegalQA dataset; (b) TrecQA dataset; (c) WikiQA dataset.

Table 6: The results of various loss functions on the model.

| Model | Dataset | | | | | |
|---|---|---|---|---|---|---|
| | LegalQA | | TrecQA | | WikiQA | |
| | MAP | MRR | MAP | MRR | MAP | MRR |
| AS-BIABC + WCE | 0.781 ± 0.002 | 0.819 ± 0.026 | 0.772 ± 0.005 | 0.780 ± 0.145 | 0.795 ± 0.010 | 0.792 ± 0.012 |
| AS-BIABC + BCE | 0.789 ± 0.000 | 0.812 ± 0.120 | 0.786 ± 0.073 | 0.804 ± 0.025 | 0.783 ± 0.074 | 0.814 ± 0.002 |
| AS-BIABC + FL | 0.842 ± 0.048 | 0.838 ± 0.056 | 0.839 ± 0.090 | 0.829 ± 0.012 | 0.832 ± 0.005 | 0.822 ± 0.006 |
| AS-BIABC + DL | 0.838 ± 0.089 | 0.808 ± 0.135 | 0.810 ± 0.074 | 0.770 ± 0.203 | 0.806 ± 0.082 | 0.804 ± 0.120 |
| AS-BIABC + TL | 0.785 ± 0.096 | 0.783 ± 0.582 | 0.821 ± 0.006 | 0.800 ± 0.041 | 0.823 ± 0.018 | 0.799 ± 0.005 |

Table 7: For the question "When were the Nobel Prize awards first given?" the table shows the top-5 answers from the model with and without reinforcement learning.

| Rank | Ranked answers w/o RL | Ranked answers by RL |
|---|---|---|
| 1 | The first *awards* ceremony took place **in 1901** | The *award* to Doctors Without Borders echoes the first Nobel Peace Prize of the century, *given* **in 1901**, of which the founder of the red cross was a corecipient |
| 2 | The five-member *awards* committee works in secrecy during its five or six meetings a year and refuses to comment on or release candidates' names | The *prizes*, first awarded **in 1901**, are always presented on Dec 10, anniversary of *Nobel*'s death |
| 3 | **In 1901**, Sweden bestowed the inaugural Nobel *Prize* in Medicine on a Berliner, Emil von Behring, for his serum against diphtheria | Among them is the winner of the first *prize* **in 1901**, Sully Prudhomme |
| 4 | The *prizes*, first awarded **in 1901**, are always presented on Dec 10, anniversary of Nobel's death | The first *awards* ceremony took place **in 1901** |
| 5 | "We all know that there are still major problems to be faced," said *awards* committee chairman Francis Sejersted | A day after the announcement, for example, critic Norman Holmes Pearson grumbled that this woman, Pearl Buck, was *given* the Nobel *Prize* in Literature |

"In 1901" is the ground truth answer, and italicized words are terms that appear in the question.

Table 8: The results of various word embeddings on the model.

| Word embedding | Dataset | | | | | |
| --- | --- | --- | --- | --- | --- | --- |
| | LegalQA | | TrecQA | | WikiQA | |
| | MAP | MRR | MAP | MRR | MAP | MRR |
| One-Hot encoding | 0.679 ± 0.042 | 0.569 ± 0.002 | 0.711 ± 0.120 | 0.653 ± 0.081 | 0.649 ± 0.089 | 0.589 ± 0.093 |
| CBOW | 0.869 ± 0.006 | 0.843 ± 0.000 | 0.889 ± 0.078 | 0.869 ± 0.120 | 0.836 ± 0.012 | 0.828 ± 0.010 |
| Skip-gram | 0.874 ± 0.052 | 0.872 ± 0.075 | 0.878 ± 0.030 | 0.858 ± 0.002 | 0.847 ± 0.014 | 0.853 ± 0.014 |
| GloVe | 0.812 ± 0.027 | 0.853 ± 0.082 | 0.795 ± 0.140 | 0.821 ± 0.074 | 0.782 ± 0.039 | 0.806 ± 0.009 |
| FastText | 0.881 ± 0.002 | 0.901 ± 0.041 | 0.886 ± 0.093 | 0.876 ± 0.002 | 0.861 ± 0.099 | 0.870 ± 0.000 |

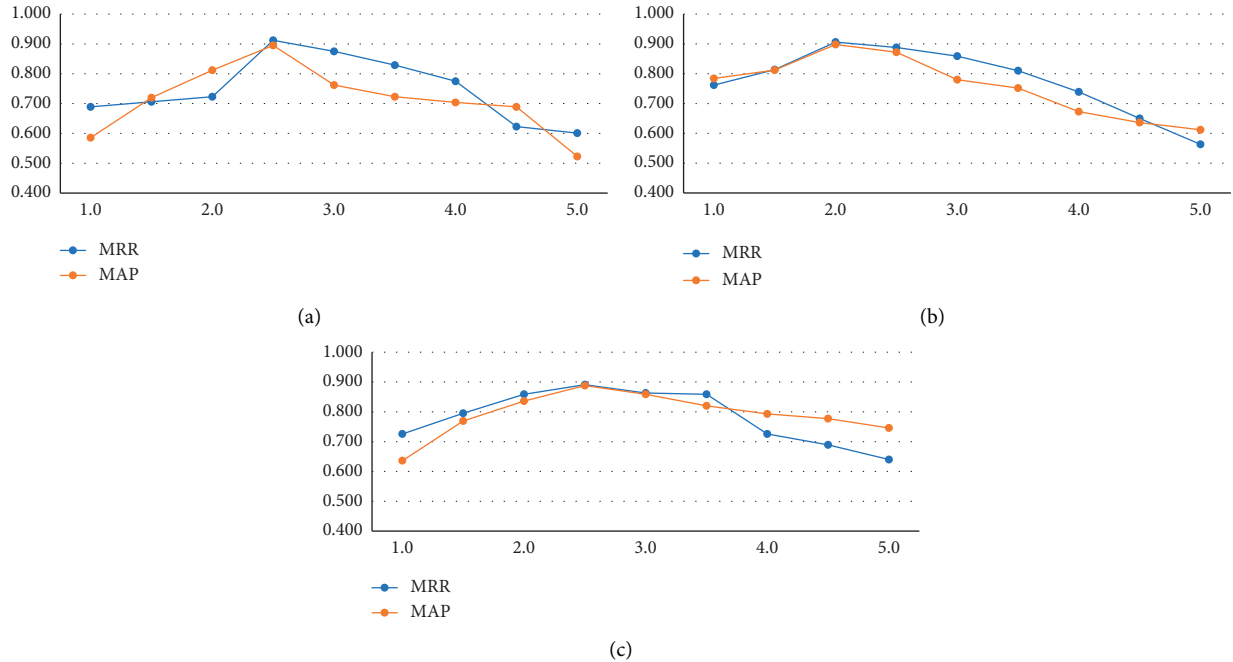

Figure 5: The process of changing the criteria by modifying the value of $F$ for the three datasets: (a) LegalQA dataset; (b) TrecQA dataset; (c) WikiQA dataset.

*5.5.4. Exploration on Word Embedding.* Word embedding is one of the main components of deep learning models because the input is interpreted as a vector, and in case of incorrect embedding, the model may be misled. This study uses the BERT model as a word embedding, developed as one of the latest embedding models. In order to check other word embeddings on our model, we employ four word embeddings: One-Hot encoding [140], CBOW [141], Skip-gram [93], GloVe [94], and FastText [95]. One-Hot encoding is the vital process of altering the categorical data variables to be supplied to deep learning algorithms, improving predictions and classification accuracy. This word embedding makes a new binary feature for each class and allocates a value of 1 to the feature of each sample that corresponds to its original class. CBOW and Skip-gram are models that use neural networks to map a word to its embedding vector. The GloVe word embedding is an unsupervised learning algorithm performed on a corpus's aggregated global word-word cooccurrence statistics. FastText is word embedding that is an extension of the Skip-gram model. Instead of learning vectors for words, this method represents each word as an $n$-gram of characters. The results of this experiment are shown in Table 8. As expected, One-Hot encoding has the worst performance among all word embeddings, so in the TrecQA dataset, where this word embedding shows the best performance, the improvement rates for the MAP and MRR criteria are about 64.70% and 72.91%, respectively. CBOW and Skip-gram perform almost identically in all three datasets due to their similar architecture, with both being superior to the GloVe word embedding. FastText serves as the best word embedding for all models but still acts poorly on the BERT model. The BERT model decreases errors by more than 11%, 10%, and 19% compared to the FastText model for the WikiQA, TrecQA, and LegalQA datasets.

*5.5.5. The Effect of the Parameter* **F** *on the Model.* To examine the effect of the parameter $F$ expressed by (10) on the proposed method algorithm performance, $F$ is set to 0.5, 1, 1.5, 2, 2.5, 3.5, 4, 4.5, and 5. The results obtained by these settings for the three datasets are shown in Figure 5. As can be seen, for the LegalQA dataset, when $F$ rises from 0 to 2, the algorithm performs better and better. However, it can be observed that when $F$ increases from 2 to 5, the method performance decreases. This means that a small or large value of $F$ weakens the algorithm performance. For the TrecQA and WikiQA datasets, the algorithm with $F$ equal to 1.5 and 2 has the best performance compared to other values.

## 6. Conclusion and Future Works

This paper presented an approach called RLAS-BIABC for AS, established on an attention mechanism-based LSTM method and the BERT word embedding, combined with an improved ABC algorithm for pretraining and reinforcement learning for training the BP algorithm. The RLAS-AM-ABC model aims to classify the two positive and negative classes, in which the positive pair includes a question and a real answer. In contrast, the negative couple carries a question and a fake answer. Due to many negative pairs in the dataset, the RLAS-BIABC is converted to an imbalanced classification. To overcome this problem, we formulate our model as a sequential decision-making process. In this regard, the environment assigned a reward to each classification act at each step, where a minority class has a higher reward. It continued until the agent mistakenly categorized a minority class sample or the number of episodes ended. Initial weighting is another essential characteristic of deep models, which can result in getting stuck in a local optimum. To solve this concern, we initialized the policy weights with the improved ABC algorithm. The paper proposed a mutual learning technique that alters the produced candidate food source with the higher fitness between two individuals chosen by a mutual learning factor. We designed experiments to examine the factors influencing the model. The analyses demonstrate the power of reinforcement learning, BERT, and the improved ABC algorithm for selecting answers.

In future work, while improving the proposed model, we will try to examine the effectiveness of the proposed classifier on other NLP applications. Another task would be to provide a model for generating the answer to a question. As a solution, we will focus on GANs, which today has many applications in almost every field, including NLP tasks.

## Data Availability

The data used to support the findings of this study are included within the article. We included the information of datasets in the articles (see part 5.1).

## Conflicts of Interest

The authors declare that they have no conflicts of interest.

## References

[1] Y. Sun, C. Chen, T. Xia, and X. Zhao, "QuGAN: quasi generative adversarial network for Tibetan question answering corpus generation," *IEEE Access*, vol. 7, pp. 116247–116255, 2019.

[2] J. Huang, "A multi-size neural network with attention mechanism for answer selection," 2021, https://arxiv.org/abs/2105.03278.

[3] W. Lei, Y. Xiang, Y. Wang, Q. Zhong, M. Liu, and M.-Y. Kan, "Linguistic properties matter for implicit discourse relation recognition: combining semantic interaction, topic continuity and attribution," in *Proceedings of the AAAI Conference on Artificial Intelligence*, vol. 32, p. 1, Louisiana, USA, February 2018.

[4] S. Li, W. Lei, Q. Wu, X. He, P. Jiang, and T.-S. Chua, "Seamlessly Unifying Attributes and Items: Conversational Recommendation for Cold-Start Users," 2020, https://arxiv.org/abs/2005.12979.

[5] S. Zhang, X. Zhang, H. Wang, J. Cheng, P. Li, and Z. Ding, "Chinese medical question answer matching using end-to-end character-level multi-scale CNNs," *Applied Sciences*, vol. 7, no. 8, p. 767, 2017.

[6] H. Yu, M. Lee, D. Kaufman et al., "Development, implementation, and a cognitive evaluation of a definitional question answering system for physicians," *Journal of Biomedical Informatics*, vol. 40, no. 3, pp. 236–251, 2007.

[7] X. Xu, L. He, A. Shimada, R.-i. Taniguchi, and H. Lu, "Learning unified binary codes for cross-modal retrieval via latent semantic hashing," *Neurocomputing*, vol. 213, pp. 191–203, 2016.

[8] X. Xu, F. Shen, Y. Yang, H. T. Shen, and X. Li, "Learning discriminative binary codes for large-scale cross-modal retrieval," *IEEE Transactions on Image Processing*, vol. 26, no. 5, pp. 2494–2507, 2017.

[9] S. Jain and T. Dodiya, "Rule based architecture for medical question answering system," in *Proceedings of the Second International Conference on Soft Computing for Problem Solving (SocProS 2012)*, pp. 1225–1233, Springer, Jaipur, India, December, 2012.

[10] X. Yao, B. Van Durme, C. Callison-Burch, and P. Clark, "Answer extraction as sequence tagging with tree edit distance," in *Proceedings of the 2013 Conference of the North American Chapter of the Association for Computational Linguistics: Human Language Technologies*, pp. 858–867, Atlanta, Georgia, June, 2013.

[11] S.-J. Yen, Y.-C. Wu, J.-C. Yang, Y.-S. Lee, C.-J. Lee, and J.-J. Liu, "A support vector machine-based context-ranking model for question answering," *Information Sciences*, vol. 224, pp. 77–87, 2013.

[12] B. L. Cairns, R. D. Nielsen, J. J. Masanz et al., "The MiPACQ clinical question answering system," in *Proceedings of the AMIA... Annual Symposium proceedings. AMIA Symposium*, vol. 2011, pp. 171–80, American Medical Informatics Association, Washington, DC, USA, October 2011.

[13] W. Lu, H. Wu, P. Jian, Y. Huang, and H. Huang, "An empirical study of classifier combination based word sense disambiguation," *IEICE - Transactions on Info and Systems*, vol. E101.D, no. 1, pp. 225–233, 2018.

[14] S. Wang and L. Cao, "Inferring implicit rules by learning explicit and hidden item dependency," *IEEE Transactions on Systems, Man, and Cybernetics: Systems*, vol. 50, no. 3, pp. 935–946, 2017.

[15] M. Heilman and N. A. Smith, "Tree edit models for recognizing textual entailments, paraphrases, and answers to questions," in *Proceedings of the Human Language Technologies: The 2010 Annual Conference of the North American Chapter of the Association for Computational Linguistics*, pp. 1011–1019, downtown Los Angeles, June 2010.

[16] H. Toba, Z.-Y. Ming, M. Adriani, and T.-S. Chua, "Discovering high quality answers in community question answering archives using a hierarchy of classifiers," *Information Sciences*, vol. 261, pp. 101–115, 2014.

[17] S. Moravvej, M. Maleki Kahaki, M. Salimi Sartakhti, and M. Joodaki, "Efficient GAN-based method for extractive summarization," *Journal of Electrical and Computer Engineering Innovations*, 2021.

[18] S. V. Moravvej, A. Mirzaei, and M. Safayani, "Biomedical text summarization using conditional generative adversarial network (CGAN)," 2021, https://arxiv.org/abs/2110.11870.

[19] M. S. Sartakhti, M. J. M. Kahaki, S. V. Moravvej, M. javadi Joortani, and A. Bagheri, "Persian language model based on BiLSTM model on COVID-19 corpus," in *Proceedings of the 2021 5th International Conference on Pattern Recognition and Image Analysis (IPRIA)*, pp. 1–5, IEEE, Kashan, Iran, April 2021.

[20] Z. Sobhaninia, H. Danesh, R. Kafieh, J. J. Balaji, and V. Lakshminarayanan, "Determination of foveal avascular zone parameters using a new location-aware deep-learning method," in *Applications of Machine Learning*, vol. 2021, p. 1184311, International Society for Optics and Photonics, 2021.

[21] Z. SobhaniNia, N. Karimi, P. Khadivi, R. Roshandel, and S. Samavi, "Brain tumor classification using medial residual encoder layers," 2020, https://arxiv.org/abs/2011.00628.

[22] Z. Sobhaninia, S. Rafiei, A. Emami et al., "Fetal ultrasound image segmentation for measuring biometric parameters using multi-task deep learning," in *Proceedings of the 2019 41st Annual International Conference of the IEEE Engineering in Medicine and Biology Society (EMBC)*, pp. 6545–6548, IEEE, Berlin, Germany, July 2019.

[23] J. Bao, D. Chen, F. Wen, H. Li, and G. Hua, "CVAE-GAN: fine-grained image generation through asymmetric training," in *Proceedings of the IEEE International Conference on Computer Vision*, pp. 2745–2754, Venice, Italy, October 2017.

[24] P. Kim, "Convolutional neural network," in *MATLAB Deep Learning*, pp. 121–147, Apress, Berkeley, CA, 2017.

[25] B. Hu, Z. Lu, H. Li, and Q. Chen, "Convolutional neural network architectures for matching natural language sentences," *Advances in Neural Information Processing Systems*, vol. 27, 2014.

[26] M. E. Peters, M. Neumann, M. Iyyer et al., "Deep contextualized word representations," 2018, https://arxiv.org/abs/1802.05365.

[27] J. Devlin, M.-W. Chang, K. Lee, and K. Toutanova, "Bert: pre-training of deep bidirectional transformers for language understanding," 2018, https://arxiv.org/abs/1810.04805.

[28] A. Vaswani, N. Shazeer, N. Parmar et al., "Attention is all you need," *Advances in Neural Information Processing Systems*, vol. 30, 2017.

[29] J. Mozafari, A. Fatemi, and M. A. Nematbakhsh, "BAS: an answer selection method using BERT language model," 2019, https://arxiv.org/abs/1911.01528.

[30] X. Xin Yao, "Evolving artificial neural networks," *Proceedings of the IEEE*, vol. 87, no. 9, pp. 1423–1447, 1999.

[31] D. E. Rumelhart, G. E. Hinton, and R. J. Williams, "Learning representations by back-propagating errors," *Nature*, vol. 323, no. 6088, pp. 533–536, 1986.

[32] C. Blum and K. Socha, "Training feed-forward neural networks with ant colony optimization: an application to pattern classification," in *Proceedings of the Fifth International Conference on Hybrid Intelligent Systems (HIS'05)*, p. 6, IEEE, Rio de Janeiro, Brazil, November 2005.

[33] R. S. Sexton, B. Alidaee, R. E. Dorsey, and J. D. Johnson, "Global optimization for artificial neural networks: a tabu search application," *European Journal of Operational Research*, vol. 106, no. 2-3, pp. 570–584, 1998.

[34] R. S. Sexton, R. E. Dorsey, and J. D. Johnson, "Optimization of neural networks: a comparative analysis of the genetic algorithm and simulated annealing," *European Journal of Operational Research*, vol. 114, no. 3, pp. 589–601, 1999.

[35] M. Dorigo, M. Birattari, and T. Stutzle, "Ant colony optimization," *IEEE Computational Intelligence Magazine*, vol. 1, no. 4, pp. 28–39, 2006.

[36] F. Glover and M. Laguna, "Tabu search," in *Handbook of Combinatorial Optimization*, pp. 2093–2229, Springer, Heidelberg, Germany, 1998.

[37] D. Bertsimas and J. Tsitsiklis, "Simulated annealing," *Statistical Science*, vol. 8, no. 1, pp. 10–15, 1993.

[38] S. Mirjalili, "Genetic algorithm, Studies in Computational Intelligence," in *Evolutionary Algorithms and Neural Networks*, pp. 43–55, Springer, Heidelberg, Germany, 2019.

[39] C. A. de Sousa, "An overview on weight initialization methods for feedforward neural networks," in *2016 International Joint Conference on Neural Networks (IJCNN)*, pp. 52–59, IEEE, Vancouver, BC, Canada, July 2016.

[40] A. Ranganathan, "The levenberg-marquardt algorithm," *Tutoral on LM algorithm*, vol. 11, no. 1, pp. 101–110, 2004.

[41] C. Ozturk and D. Karaboga, "Hybrid artificial bee colony algorithm for neural network training," in *Proceedings of the 2011 IEEE congress of Evolutionary Computation (CEC)*, pp. 84–88, IEEE, New Orleans, LA, USA, July 2011.

[42] S. V. Moravvej, S. J. Mousavirad, M. H. Moghadam, and M. Saadatmand, "An LSTM-Based Plagiarism Detection via Attention Mechanism and a Population-Based Approach for Pre-training Parameters with Imbalanced Classes," *Neural Information Processing*, Springer International Publishing, vol. 13110, , pp. 690–701, 2021.

[43] Z. Abdi Alkareem Alyasseri, O. A. Alomari, M. A. Al-Betar et al., "EEG channel selection using multiobjective cuckoo search for person identification as protection system in healthcare applications," *Computational Intelligence and Neuroscience*, vol. 2022, Article ID 5974634, 18 pages, 2022.

[44] S. Vakilian, S. V. Moravvej, and A. Fanian, "Using the cuckoo algorithm to optimizing the response time and energy consumption cost of fog nodes by considering collaboration in the fog layer," in *Proceedings of the 2021 5th International Conference on Internet of Things and Applications (IoT)*, pp. 1–5, IEEE, University of Isfahan, Iran, May 2021.

[45] S. Vakilian, S. V. Moravvej, and A. Fanian, "Using the artificial bee colony (ABC) algorithm in collaboration with the fog nodes in the Internet of Things three-layer architecture," in *Proceedings of the 2021 29th Iranian Conference on Electrical Engineering (ICEE)*, pp. 509–513, IEEE, Tehran, Iran, May 2021.

[46] S. J. Mousavirad, G. Schaefer, I. Korovin, D. Oliva, and Rde-Op, "RDE-OP: a region-based differential evolution algorithm incorporation opposition-based learning for optimising the learning process of multi-layer neural networks," in *Proceedings of the International Conference on the Applications of Evolutionary Computation (Part of EvoStar)*, pp. 407–420, Springer, Seville, Spain, April 2021.

[47] H. Han, W.-Y. Wang, B.-H. Mao, and Borderline-Smote, "Borderline-SMOTE: a new over-sampling method in imbalanced data sets learning," in *Proceedings of the International Conference on Intelligent Computing*, pp. 878–887, Springer, Hefei, China, August 2005.

[48] I. Mani and I. Zhang, "kNN approach to unbalanced data distributions: a case study involving information extraction,"

in *Proceedings of the workshop on learning from imbalanced datasets*, vol. 126, ICML United States, Washington, DC, August 2003.

[49] S. Wang, W. Liu, J. Wu, L. Cao, Q. Meng, and P. J. Kennedy, "Training deep neural networks on imbalanced data sets," in *Proceedings of the 2016 International Joint Conference on Neural Networks (IJCNN)*, pp. 4368–4374, IEEE, Vancouver, BC, Canada, July 2016.

[50] C. Huang, Y. Li, C. C. Loy, and X. Tang, "Learning deep representation for imbalanced classification," in *Proceedings of the IEEE Conference on Computer Vision and Pattern Recognition*, pp. 5375–5384, Las Vegas, Nevada, July 2016.

[51] Y. Yan, M. Chen, M.-L. Shyu, and S.-C. Chen, "Deep learning for imbalanced multimedia data classification," in *Proceedings of the 2015 IEEE International Symposium on Multimedia (ISM)*, pp. 483–488, IEEE, Miami, FL, USA, December 2015.

[52] S. H. Khan, M. Hayat, M. Bennamoun, F. A. Sohel, and R. Togneri, "Cost-sensitive learning of deep feature representations from imbalanced data," *IEEE Transactions on Neural Networks and Learning Systems*, vol. 29, no. 8, pp. 3573–3587, 2017.

[53] Q. Dong, S. Gong, and X. Zhu, "Imbalanced deep learning by minority class incremental rectification," *IEEE Transactions on Pattern Analysis and Machine Intelligence*, vol. 41, no. 6, pp. 1367–1381, 2018.

[54] Q. Dong, S. Gong, and X. Zhu, "Class rectification hard mining for imbalanced deep learning," in *Proceedings of the IEEE International Conference on Computer Vision*, pp. 1851–1860, Venice, Italy, October 2017.

[55] V. Mnih, K. Kavukcuoglu, D. Silver et al., "Playing Atari with Deep Reinforcement Learning," 2013, https://arxiv.org/abs/1312.5602.

[56] S. Gu, E. Holly, T. Lillicrap, and S. Levine, "Deep reinforcement learning for robotic manipulation with asynchronous off-policy updates," in *Proceedings of the 2017 IEEE International Conference on Robotics and Automation (ICRA)*, pp. 3389–3396, IEEE, Singapore, May 2017.

[57] X. Zhao, L. Zhang, L. Xia, Z. Ding, D. Yin, and J. Tang, "Deep reinforcement learning for list-wise recommendations," 2017, https://arxiv.org/abs/1801.00209.

[58] S. Wan, M. Dras, R. Dale, and C. Paris, "Using dependency-based features to take the'para-farce'out of paraphrase," in *Proceedings of the Australasian language technology workshop*, pp. 131–138, Sydney, Australia, November 2006.

[59] G. A. Miller and WordNet, *An Electronic Lexical Database*, MIT press, Cambridge, MA, 1998.

[60] V. Punyakanok, D. Roth, and W.-t. Yih, "Natural Language Inference via Dependency Tree Mapping: An Application to Question Answering," 2004, https://www.ideals.illinois.edu/handle/2142/11100.

[61] M. Wang and C. D. Manning, "Probabilistic tree-edit models with structured latent variables for textual entailment and question answering," in *Proceedings of the 23rd International Conference on Computational Linguistics (Coling 2010)*, pp. 1164–1172, Beijing, China, August 2010.

[62] S. W.-t. Yih, M.-W. Chang, C. Meek, and A. Pastusiak, "Question Answering Using Enhanced Lexical Semantic Models," in *Proceedings of the 51st Annual Meeting of the Association for Computational Linguistics*, Sofia, Bulgaria's capital, August 2013.

[63] J. M. Giorgi and G. D. Bader, "Towards reliable named entity recognition in the biomedical domain," *Bioinformatics*, vol. 36, no. 1, pp. 280–286, 2020.

[64] A. Severyn and A. Moschitti, "Automatic feature engineering for answer selection and extraction," in *Proceedings of the 2013 Conference on Empirical Methods in Natural Language Processing*, pp. 458–467, Seattle, Washington, USA, October 2013.

[65] S. Wang and J. Jiang, "A compare-aggregate model for matching text sequences," 2016, https://arxiv.org/abs/1611.01747.

[66] S. Yoon, F. Dernoncourt, D. S. Kim, T. Bui, and K. Jung, "A compare-aggregate model with latent clustering for answer selection," in *Proceedings of the 28th ACM International Conference on Information and Knowledge Management*, pp. 2093–2096, Beijing, China, November 2019.

[67] A. Severyn and A. Moschitti, "Learning to rank short text pairs with convolutional deep neural networks," in *Proceedings of the 38th International ACM SIGIR Conference on Research and Development in Information Retrieval*, pp. 373–382, Santiago Chile, August 2015.

[68] D. Wang and E. Nyberg, "A long short-term memory model for answer sentence selection in question answering," in *Proceedings of the 53rd Annual Meeting of the Association for Computational Linguistics and the 7th International Joint Conference on Natural Language Processing*, vol. 2, pp. 707–712, Beijing, China, July 2015.

[69] J. Bromley, J. W. Bentz, L. Bottou et al., "Signature verification using a "siamese" time delay neural network," *International Journal of Pattern Recognition and Artificial Intelligence*, vol. 07, no. 04, pp. 669–688, 1993.

[70] L. Yu, K. M. Hermann, P. Blunsom, and S. Pulman, "Deep learning for answer sentence selection," 2014, https://arxiv.org/abs/1412.1632.

[71] M. Feng, B. Xiang, M. R. Glass, L. Wang, and B. Zhou, "Applying deep learning to answer selection: a study and an open task," in *Proceedings of the 2015 IEEE Workshop on Automatic Speech Recognition and Understanding (ASRU)*, pp. 813–820, IEEE, Scottsdale, Arizona, USA, December 2015.

[72] H. He, K. Gimpel, and J. Lin, "Multi-perspective sentence similarity modeling with convolutional neural networks," in *Proceedings of the 2015 Conference on Empirical Methods in Natural Language Processing*, pp. 1576–1586, Lisbon, Portugal, September 2015.

[73] J. Rao, H. He, and J. Lin, "Noise-contrastive estimation for answer selection with deep neural networks," in *Proceedings of the 25th ACM International on Conference on Information and Knowledge Management*, pp. 1913–1916, Indianapolis Indiana, USA, October 2016.

[74] H. T. Madabushi, M. Lee, and J. Barnden, "Integrating question classification and deep learning for improved answer selection," in *Proceedings of the 27th International Conference on Computational Linguistics*, pp. 3283–3294, Santa Fe, New-Mexico, August 2018.

[75] S. Kamath, B. Grau, and Y. Ma, "Predicting and integrating expected answer types into a simple recurrent neural network model for answer sentence selection," *Computación Y Sistemas*, vol. 23, no. 3, 2019.

[76] S. Kumar, K. Dixit, and K. Shah, "Interpreting text classifiers by learning context-sensitive influence of words," in *Proceedings of the First Workshop on Trustworthy Natural Language Processing*, pp. 55–67, Online, June 2021.

[77] T. Lai, T. Bui, and S. Li, "A review on deep learning techniques applied to answer selection," in *Proceedings of the 27th International Conference on Computational Linguistics*, pp. 2132–2144, Santa Fe, New-Mexico, August 2018.

[78] Y. Zhang and Y. Peng, "Research on answer selection based on LSTM," in *Proceedings of the 2018 International Conference on Asian Language Processing (IALP)*, pp. 357–361, IEEE, Bandung, Indonesia, November 2018.

[79] L. Yang, Q. Ai, J. Guo, and W. B. Croft, "aNMM: ranking short answer texts with attention-based neural matching model," in *Proceedings of the 25th ACM International on Conference on Information and Knowledge Management*, pp. 287–296, Indianapolis Indiana, USA, October 2016.

[80] D. Bahdanau, K. Cho, and Y. Bengio, "Neural machine translation by jointly learning to align and translate," 2014, https://arxiv.org/abs/1409.0473.

[81] G. Shen, Y. Yang, and Z.-H. Deng, "Inter-weighted alignment network for sentence pair modeling," in *Proceedings of the 2017 Conference on Empirical Methods in Natural Language Processing*, pp. 1179–1189, Copenhagen, Denmark, September 2017.

[82] Y. Tay, L. A. Tuan, and S. C. Hui, "Co-stack residual affinity networks with multi-level attention refinement for matching text sequences," 2018, https://arxiv.org/abs/1810.02938.

[83] H. He, J. Wieting, K. Gimpel, J. Rao, and J. Lin, "UMD-TTIC-UW at SemEval-2016 Task 1: attention-based multi-perspective convolutional neural networks for textual similarity measurement," in *Proceedings of the 10th International Workshop on Semantic Evaluation (SemEval-2016)*, pp. 1103–1108, San Diego, California, USA, June 2016.

[84] L. R. Medsker and L. Jain, "Recurrent neural networks," *Design and Applications*, vol. 5, pp. 64–67, 2001.

[85] S. Abujar, A. K. M. Masum, S. M. H. Chowdhury, M. Hasan, and S. A. Hossain, "Bengali text generation using bi-directional RNN," in *2019 10th International Conference on Computing, Communication and Networking Technologies (ICCCNT)*, pp. 1–5, IEEE, IIT, Kanpur, July 2019.

[86] A. K. M. Masum, S. Abujar, M. A. I. Talukder, A. S. A. Rabby, and S. A. Hossain, "Abstractive method of text summarization with sequence to sequence RNNs," pp. 1–5, IEEE.

[87] Y. Yang, J. Zhou, J. Ai et al., "Video captioning by adversarial LSTM," *IEEE Transactions on Image Processing*, vol. 27, no. 11, pp. 5600–5611, 2018.

[88] S. Hochreiter and J. Schmidhuber, "Long short-term memory," *Neural Computation*, vol. 9, no. 8, pp. 1735–1780, 1997.

[89] A. Graves, "Generating sequences with recurrent neural networks," 2013, https://arxiv.org/abs/1308.0850.

[90] S. V. Moravvej, M. Joodaki, M. J. M. Kahaki, and M. S. Sartakhti, "A method based on an attention mechanism to measure the similarity of two sentences," in *Proceedings of the 2021 7th International Conference on Web Research (ICWR)*, pp. 238–242, IEEE, Tehran, Iran, May 2021.

[91] A. Graves and J. Schmidhuber, "Framewise phoneme classification with bidirectional LSTM and other neural network architectures," *Neural Networks : The Official Journal of the International Neural Network Society*, vol. 18, no. 5-6, pp. 602–610, 2005.

[92] J.-Á. González, E. Segarra, F. García-Granada, E. Sanchis, and L. r. 1.-F. Hurtado, "Siamese hierarchical attention networks for extractive summarization," *Journal of Intelligent and Fuzzy Systems*, vol. 36, no. 5, pp. 4599–4607, 2019.

[93] C. McCormick, "Word2vec tutorial-the skip-gram model," 2016, http://mccormickml.com/2016/04/19/word2vec-tutorial-the-skip-gram-model.

[94] J. Pennington, R. Socher, and C. D. Manning, "Glove: global vectors for word representation," in *Proceedings of the 2014 Conference on Empirical Methods in Natural Language Processing (EMNLP)*, pp. 1532–1543, Doha, Qatar, October 2014.

[95] S. Thavareesan and S. Mahesan, "Sentiment lexicon expansion using Word2vec and fastText for sentiment prediction in Tamil texts," in *Proceedings of the 2020 Moratuwa Engineering Research Conference (MERCon)*, pp. 272–276, IEEE, Moratuwa, Sri Lanka, July 2020.

[96] H. S. Alatawi, A. M. Alhothali, and K. M. Moria, "Detecting white supremacist hate speech using domain specific word embedding with deep learning and BERT," *IEEE Access*, vol. 9, pp. 106363–106374, 2021.

[97] J. Devlin and M.-W. Chang, "Open sourcing BERT: state-of-the-art pre-training for natural language processing," *Google AI Blog*, vol. 2, 2018.

[98] K. Arulkumaran, M. P. Deisenroth, M. Brundage, and A. A. Bharath, "Deep reinforcement learning: a brief survey," *IEEE Signal Processing Magazine*, vol. 34, no. 6, pp. 26–38, 2017.

[99] S. A. H. Minoofam, A. Bastanfard, and M. R. Keyvanpour, "TRCLA: a transfer learning approach to reduce negative transfer for cellular learning automata," *IEEE Transactions on Neural Networks and Learning Systems*, 2021.

[100] Z. Movahedi and A. Bastanfard, "Toward Competitive Multi-Agents in Polo Game Based on Reinforcement learning," *Multimedia Tools and Applications*, vol. 80, no. 17, pp. 26773–26793, 2021.

[101] M. A. Wiering, H. Van Hasselt, A.-D. Pietersma, and L. Schomaker, "Reinforcement learning algorithms for solving classification problems," in *Proceedings of the 2011 IEEE Symposium on Adaptive Dynamic Programming and Reinforcement Learning (ADPRL)*, pp. 91–96, IEEE, Paris, France, April 2011.

[102] J. Feng, M. Huang, L. Zhao, Y. Yang, and X. Zhu, "Reinforcement learning for relation classification from noisy data," in *Proceedings of the AAAI Conference on Artificial Intelligence*, vol. 32, p. 1, New Orleans, Louisiana, USA, February 2018.

[103] T. Zhang, M. Huang, and L. Zhao, "Learning structured representation for text classification via reinforcement learning," in *Proceedings of the Thirty-Second AAAI Conference on Artificial Intelligence*, New Orleans, Louisiana, USA, February 2018.

[104] D. Liu and T. Jiang, "Deep reinforcement learning for surgical gesture segmentation and classification," *Medical Image Computing and Computer Assisted Intervention - MICCAI 2018*, in *Proceedings of the International Conference on Medical Image Computing and Computer-Assisted Intervention*, pp. 247–255, Springer, September 2018.

[105] D. Zhao, Y. Chen, and L. Lv, "Deep reinforcement learning with visual attention for vehicle classification," *IEEE Transactions on Cognitive and Developmental Systems*, vol. 9, no. 4, pp. 356–367, 2016.

[106] J. Janisch, T. Pevný, and V. Lisý, "Classification with costly features using deep reinforcement learning," *Proceedings of the AAAI Conference on Artificial Intelligence*, vol. 33, no. 01, pp. 3959–3966, 2019.

[107] C. Martinez, G. Perrin, E. Ramasso, and M. Rombaut, "A deep reinforcement learning approach for early classification of time series," in *Proceedings of the 2018 26th European Signal Processing Conference (EUSIPCO)*, pp. 2030–2034, IEEE, Rome, Italy, September 2018.

[108] L. Abdi and S. Hashemi, "An ensemble pruning approach based on reinforcement learning in presence of multi-class imbalanced data," *Advances in Intelligent Systems and*

Computing, in *Proceedings of the Third International Conference on Soft Computing for Problem Solving*, pp. 589–600, Springer, Roorkee, India, December 2013.
[109] A. K. Dixit and J. J. Sherrerd, *Optimization in Economic Theory*, Oxford University Press on Demand, Oxford, England, 1990.
[110] E. M. Voorhees, "The evaluation of question answering systems: lessons learned from the TREC QA track," in *Question Answering: Strategy and Resources Workshop Program*, p. 6, Citeseer, New Jersey, USA, 2002.
[111] W. Huang, J. Jiang, Q. Qu, and M. Yang, "AILA: A Question Answering System in the Legal Domain," in *Proceedings of the Twenty-Ninth International Conference on International Joint Conferences on Artificial Intelligence*, pp. 5258–5260, Yokohama, July 2020.
[112] Y. Yang, W.-t. Yih, and C. Meek, "Wikiqa: a challenge dataset for open-domain question answering," in *Proceedings of the 2015 Conference on Empirical Methods in Natural Language Processing*, pp. 2013–2018, Lisbon, Portugal, September 2015.
[113] Y. Shen, Y. Deng, M. Yang et al., "Knowledge-aware attentive neural network for ranking question answer pairs," in *Proceedings of the The 41st International ACM SIGIR Conference on Research & Development in Information Retrieval*, pp. 901–904, Ann Arbor, Michigan, July 2018.
[114] S. V. Moravvej, M. J. M. Kahaki, M. S. Sartakhti, and A. Mirzaei, "A method based on attention mechanism using bidirectional long-short term memory (BLSTM) for question answering," in *Proceedings of the 2021 29th Iranian Conference on Electrical Engineering (ICEE)*, pp. 460–464, IEEE, Tehran, Iran, Islamic Republic of, May 2021.
[115] Y. Matsubara, T. Vu, and A. Moschitti, "Reranking for efficient transformer-based answer selection," in *Proceedings of the 43rd International ACM SIGIR Conference on Research and Development in Information Retrieval*, pp. 1577–1580, Xi'an, China, July 2020.
[116] S. Kim, I. Kang, and N. Kwak, "Semantic sentence matching with densely-connected recurrent and co-attentive information," *Proceedings of the AAAI Conference on Artificial Intelligence*, vol. 33, no. 01, pp. 6586–6593, 2019.
[117] Y. Song, Q. V. Hu, and L. He, "P-CNN: e," *Knowledge-Based Systems*, vol. 169, pp. 67–79, 2019.
[118] G. Bao, Y. Wei, X. Sun, and H. Zhang, "Double attention recurrent convolution neural network for answer selection," *Royal Society Open Science*, vol. 7, no. 5, p. 191517, 2020.
[119] Q. Wang, W. Wu, Y. Qi, and Z. Xin, "Combination of active learning and self-paced learning for deep answer selection with bayesian neural network," in *Proceedings of the ECAI 2020*, pp. 1587–1594, IOS Press, Santiago de Compostela, Spain, July 2020.
[120] W. Huang, Q. Qu, and M. Yang, "Interactive knowledge-enhanced attention network for answer selection," *Neural Computing and Applications*, vol. 32, no. 15, pp. 11343–11359, 2020.
[121] V. V. Phansalkar and P. S. Sastry, "Analysis of the back-propagation algorithm with momentum," *IEEE Transactions on Neural Networks*, vol. 5, no. 3, pp. 505–506, 1994.
[122] M. Hagan, H. Demuth, and M. Beale, *Neural Network Design*, PWS Publishing Co, Boston, MA, 1996.
[123] C.-C. Yu and B.-D. Liu, "A backpropagation algorithm with adaptive learning rate and momentum coefficient," in *Proceedings of the 2002 International Joint Conference on Neural Networks*, vol. 2, pp. 1218–1223, IEEE, Honolulu, HI, USA, May 2002.
[124] R. Battiti, "First- and second-order methods for learning: between steepest descent and Newton's method," *Neural Computation*, vol. 4, no. 2, pp. 141–166, 1992.
[125] F. D. Foresee and M. T. Hagan, "Gauss-Newton approximation to Bayesian learning," in *Proceedings of the international conference on neural networks (ICNN'97)*, vol. 3, pp. 1930–1935, IEEE, Houston, TX, USA, June 1997.
[126] S. Mirjalili, S. M. Mirjalili, and A. Lewis, "Grey wolf optimizer," *Advances in Engineering Software*, vol. 69, pp. 46–61, 2014.
[127] X.-S. Yang, "A new metaheuristic bat-inspired algorithm," in *Proceedings of the Nature Inspired Cooperative Strategies for Optimization (NICSO 2010)*, pp. 65–74, Springer, Granada, Spain, May 2010.
[128] S. Mirjalili, "Dragonfly algorithm: a new meta-heuristic optimization technique for solving single-objective, discrete, and multi-objective problems," *Neural Computing & Applications*, vol. 27, no. 4, pp. 1053–1073, 2016.
[129] D. Bairathi and D. Gopalani, "Salp swarm algorithm (SSA) for training feed-forward neural networks," in *Soft Computing for Problem Solving*, pp. 521–534, Springer, Singapore, 2019.
[130] X.-S. Yang and S. Deb, "Cuckoo search via Lévy flights," in *Proceedings of the 2009 World congress on Nature & Biologically Inspired Computing (NaBIC)*, pp. 210–214, IEEE, Coimbatore, India, December 2009.
[131] S. J. Mousavirad and H. Ebrahimpour-Komleh, "Human mental search: a new population-based metaheuristic optimization algorithm," *Applied Intelligence*, vol. 47, no. 3, pp. 850–887, 2017.
[132] S. Mirjalili and A. Lewis, "The whale optimization algorithm," *Advances in Engineering Software*, vol. 95, pp. 51–67, 2016.
[133] D. Karaboga and B. Basturk, "A powerful and efficient algorithm for numerical function optimization: artificial bee colony (ABC) algorithm," *Journal of Global Optimization*, vol. 39, no. 3, pp. 459–471, 2007.
[134] V. Pihur, S. Datta, and S. Datta, "Weighted rank aggregation of cluster validation measures: a Monte Carlo cross-entropy approach," *Bioinformatics*, vol. 23, no. 13, pp. 1607–1615, 2007.
[135] S. Xie and Z. Tu, "Holistically-nested edge detection," in *Proceedings of the IEEE International Conference on Computer Vision*, pp. 1395–1403, NW Washington, DC, USA, December 2015.
[136] T.-Y. Lin, P. Goyal, R. Girshick, K. He, and P. Dollár, "Focal loss for dense object detection," in *Proceedings of the IEEE International Conference on Computer Vision*, pp. 2980–2988, Venice, Italy, October 2017.
[137] C. H. Sudre, W. Li, T. Vercauteren, S. Ourselin, and M. Jorge Cardoso, "Generalised dice overlap as a deep learning loss function for highly unbalanced segmentations," in *Deep Learning in Medical Image Analysis and Multimodal Learning for Clinical Decision Support*, pp. 240–248, Springer, Heidelberg, Germany, 2017.
[138] S. S. M. Salehi, D. Erdogmus, and A. Gholipour, "Tversky loss function for image segmentation using 3D fully convolutional deep networks," *Machine Learning in Medical Imaging*, in *Proceedings of the International Workshop on*

*Machine Learning in Medical Imaging*, pp. 379–387, Quebec, Canada, Springer, September 2017.
[139] S. Jadon, "A survey of loss functions for semantic segmentation," in *Proceedings of the 2020 IEEE Conference on Computational Intelligence in Bioinformatics and Computational Biology (CIBCB)*, pp. 1–7, IEEE, Via del Mar, Chile, October 2020.
[140] G. Hackeling, *Mastering Machine Learning with Scikit-Learn*, Packt Publishing, Birmingham, UK, 2014.
[141] S. Sonkar, A. E. Waters, and R. G. Baraniuk, "Attention word embedding," 2020, https://arxiv.org/abs/2006.00988.